\documentclass[letterpaper, 10 pt, conference]{ieeeconf}  

\IEEEoverridecommandlockouts                              

\usepackage{cite}
\usepackage{amsmath,amssymb,amsfonts}
\usepackage{graphicx}
\usepackage{textcomp}
\usepackage{xcolor}

\def\BibTeX{{\rm B\kern-.05em{\sc i\kern-.025em b}\kern-.08em
    T\kern-.1667em\lower.7ex\hbox{E}\kern-.125emX}}

\usepackage{algorithm}
\usepackage{algorithmicx}  
\usepackage[noend]{algpseudocode}
\usepackage{amsfonts}
\usepackage{amsmath}
\usepackage{amsthm}
\usepackage{bm}
\usepackage{booktabs}
\usepackage{diagbox}
\usepackage{dsfont} 
\let\labelindent\relax  
\usepackage{enumitem}
\usepackage{etoolbox}
\usepackage{graphicx}  
\usepackage{hyperref}  
\usepackage{lipsum}  
\usepackage{mathtools}
\usepackage{multirow}
\usepackage{nicefrac}
\usepackage[final]{pdfpages}
\usepackage{siunitx}
\usepackage[caption=false]{subfig}
\usepackage{tabu}
\usepackage{threeparttable}  
\usepackage{tikz}
\usepackage[textsize=small]{todonotes}
\usepackage{xcolor}
\usepackage{xfrac}
\usepackage{blindtext}
\usepackage{placeins}

\usepackage{pgfplots}   
\pgfplotsset{compat=newest}
\usepgfplotslibrary{groupplots}
\usepgfplotslibrary{fillbetween}
\usetikzlibrary{pgfplots.statistics} 
\usetikzlibrary{arrows.meta}

\usepackage{mathabx}   

\usepackage{marvosym}  
\graphicspath{{img/}}
\hypersetup{colorlinks=false}

\definecolor{githubColor}{HTML}{2EA44F}

\definecolor{newGray}{HTML}{808080}

\definecolor{backGray}{rgb}{0.3, 0.3, 0.3}%
\definecolor{matlabYellow}{rgb}{0.9290, 0.6940, 0.1250}%
\definecolor{matlabPurple}{rgb}{0.4940, 0.1840, 0.5560}%
\definecolor{matlabLBlue}{rgb}{0.3010, 0.7450, 0.9330}%
\definecolor{matlabGreen}{rgb}{0.4660, 0.6740, 0.1880}%
\definecolor{matlabRed}{rgb}{0.8500, 0.3250, 0.0980}%
\definecolor{matlabBlue}{rgb}{0, 0.4470, 0.7410}%
\definecolor{matlabDarkRed}{rgb}{0.6350 0.0780 0.1840}%

\definecolor{TAB_RED}{RGB}{214,39,40}
\definecolor{TAB_BLUE}{RGB}{31, 119, 180}
\definecolor{TAB_ORANGE}{RGB}{255, 127, 14}

\definecolor{colorCircle}{HTML}{0072BD}

\definecolor{colorRect}{HTML}{D95319}

\newcolumntype{O}[1]{S[detect-weight, mode=text, table-format=#1]}

\renewcommand{\bfseries}{\fontseries{b}\selectfont} 
\robustify\bfseries             
\newrobustcmd{\B}{\bfseries}

\newcommand\copyrighttext{\footnotesize \textcopyright~2024 IEEE. Personal use of this material is permitted. Permission from IEEE must be obtained for all other uses, in any current or future media, including reprinting/republishing this material for advertising or promotional purposes, creating new collective works, for resale or redistribution to servers or lists, or reuse of any copyrighted component of this work in other works.
}

\newcommand\copyrightnotice{%
    \begin{tikzpicture}[remember picture,overlay]%
 	\node[anchor=south, xshift=-0pt, yshift=20pt] at (current page.south)%
 	{\fbox{\parbox{\dimexpr\textwidth-\fboxsep-\fboxrule\relax}{\copyrighttext}}};%
 	\end{tikzpicture}%
}

\newtheoremstyle{tstyle}
  {}
  {}
  {\itshape}
  {}
  {\bfseries}
  {.}
  { }
  {\thmname{#1}\thmnumber{ #2}\thmnote{ (#3)}}%
\theoremstyle{tstyle}

\renewcommand{\vec}[1]{\boldsymbol{#1}}

\newcommand{\pspace}{\,}  

\newtheorem{definition}{Definition}

\title{%
\LARGE \bf Self-Assessment of Evidential Grid Map Fusion\\for Robust Motion Planning%
}

\author{Oliver Schumann, Thomas Wodtko, Michael Buchholz, and Klaus Dietmayer%
\thanks{This work was supported by the State Ministry of Economic Affairs, Labour and Tourism Baden-Württemberg (project U-Shift\,II, AZ\,3-433.62-DLR/60).}%
\thanks{All authors are with the Institute of Measurement, Control and Microtechnology, Ulm University, Albert-Einstein-Allee 41, 89081 Ulm, Germany {\tt\footnotesize \{firstname\}.\{lastname\}@uni-ulm.de}}%
}

\begin{document}

\maketitle
\copyrightnotice%
\begin{abstract}
Conflicting sensor measurements pose a huge problem for the environment representation of an autonomous robot.
Therefore, in this paper, we address the self-assessment of an evidential grid map in which data from conflicting LiDAR sensor measurements are fused, followed by methods for robust motion planning under these circumstances.
First, conflicting measurements aggregated in Subjective-Logic-based evidential grid maps are classified. Then, a self-assessment framework evaluates these conflicts and estimates their severity for the overall system by calculating a degradation score. This enables the detection of calibration errors and insufficient sensor setups.
In contrast to other motion planning approaches, the information gained from the evidential grid maps is further used inside our proposed path-planning algorithm. Here, the impact of conflicting measurements on the current motion plan is evaluated, and a robust and curious path-planning strategy is derived to plan paths under the influence of conflicting data.
This ensures that the system integrity is maintained in severely degraded environment representations which can prevent the unnecessary abortion of planning tasks. 
\end{abstract}


\FloatBarrier
\section{Introduction}
\label{sec:intro}
\begin{figure}[t]
    \centering
    \includegraphics[width=1.0\columnwidth]{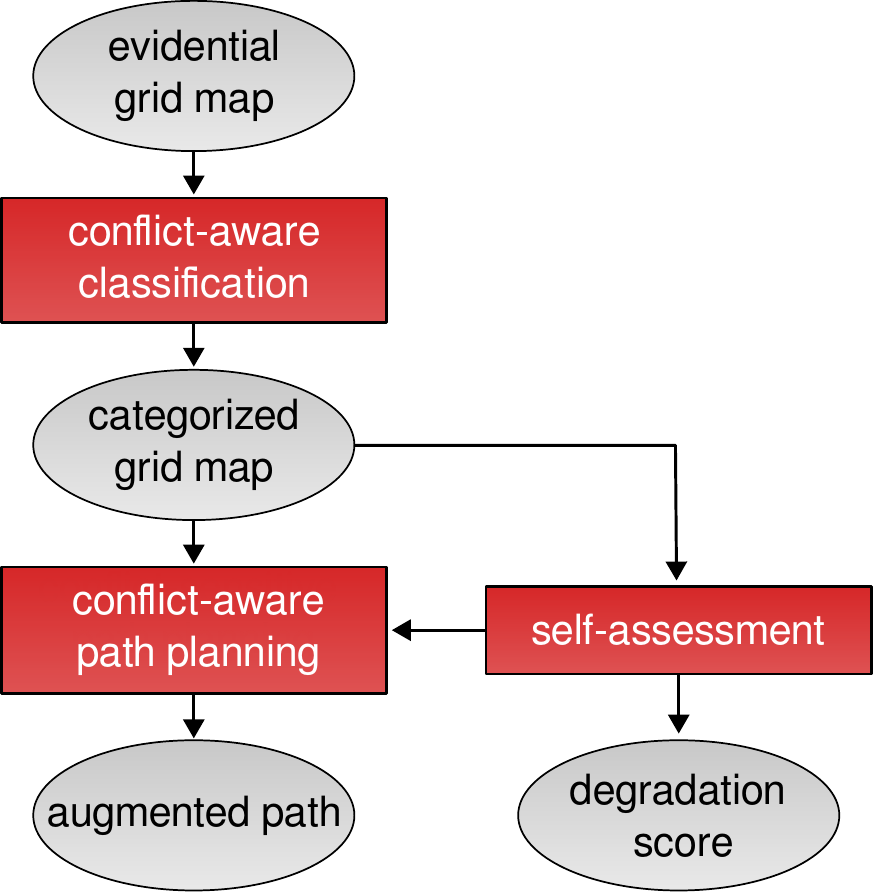}
    \caption{%
        Sensor measurements are fused in the evidential grid map whose cells are then categorized in the conflict-aware classification module. 
        They are evaluated in the self-assessment module and assigned movement costs or set to non-drivable depending on their distance to the ego vehicle.
        This leads to conflict-aware path planning on the categorized grid map, which retains information about conflicting poses in the path for underlying stages of motion planning.
        Last, the categorized grid map is used to calculate a degradation score, which evaluates the overall integrity of the sensor data.
    }
    \label{fig:overview}
    \vspace{-0.2cm}
\end{figure}
In the fields of robotics and automated driving, a robust and correctly working sensor setup is crucial for the perception of the environment. 
In this paper, we investigate how conflicting sensor data fused in evidential grid maps can be detected, analyze their impact on the overall system, and evaluate to what extent this information can be used in the underlying motion planning stage.

Possible errors in the perception stage can be complete sensor failures or partial performance degradation. 
While the former is trivial to detect, the latter is more dangerous, as it can generate misleading sensor data. 
The reasons for the latter errors are manifold. 
For example, sensors could be calibrated incorrectly during an inspection or could slowly shift or rotate due to vibrations. 
Sensor data could suffer from increased latency caused by high system load, or sensors could even be damaged during an accident or be manipulated deliberately. 
In these cases, the measurements of the malfunctioning sensor will not match the ones from the other sensors, which can lead to conflicting measurements. 

However, an automated vehicle (AV) must be able to reach a safe state even if these errors distort its perceived environmental model.
Further, there are scenarios where even complicated tasks must be executed to reach a safe state.
For example, errors in occluded parts of a curve can lead to danger for the vehicle and other traffic participants if the vehicle remains in this part of the road.
Lastly, there are cases where it is completely impossible to access an autonomous robot, for example, errors in unmanned rovers or autonomous search robots in catastrophe areas.

To summarize, an AV must detect these errors in the perceptions stage, estimate their impact on the mission, and react as effectively as possible. 
Hence, the main contributions of this work are visualized in Fig.~\ref{fig:overview}, which are:
\begin{itemize}
    \item a Subjective-Logic-based classification of grid cells as conflicting if so,
    \item a self-assessment framework to evaluate the impact of conflicting data; and
    \item a conflict-aware path planning concept to plan successfully within degraded environment representations.
\end{itemize}
Therefore, this paper is structured as follows: 
First, related work is summarized in Sec.~\ref{sec:related-work}, followed by the foundations in Sec.~\ref{sec:foundations}.
After this, Sec.~\ref{sec:problem_formulation} presents an exemplary error of a sensor setup within the Carla simulator \cite{carla2023new}. 
Then, the used methods of conflict-aware classification, self-assessment, and conflict-aware path planning are presented in Sec.~\ref{sec:method} followed by the evaluation in Sec.~\ref{sec:eval}. Finally, the content is summarized in Sec.~\ref{sec:conclusion}.

\FloatBarrier
\section{Related Work}\label{sec:related-work}
The first topic relevant to this work is the handling of sensor errors.
It is extensively described by means of anomaly detection in \cite{VanWyk2020Real-timeVehicles, Wang2021Real-TimeSensors}. 
Here, outliers are detected using Deep Neural Networks and Kalman filters. 

If the data from multiple sensors should be fused, evidential grid mapping approaches using the Dempster-Shafer Theory (DST)~\cite{dempster1967upper,Shafer1976} were increasingly used as in~\cite{Tanzmeister2017EvidentialMapping, Steyer2018Grid-basedTracking, nuss2018, wodtko2023adaptive}.
With evidential logic, data from multi-sensor setups with different sensor modalities and capabilities can be fused.

In contrast to grid mapping approaches using Bayesian probabilities, they also have the advantage of being capable of handling conflicting measurements~\cite{thrun2005probabilistic, josang2016}. This topic of fusing conflicting measurements was investigated, for example, for conflicting sonar data~\cite{Lee2009EffectiveSensors}, and for an improved fusion of heterogenous sensor setups in~\cite{richter2022}.

Further, the authors of~\cite{griebelNew2020, griebelNew2022, griebelMulti} handled conflicts between incoming measurements and the corresponding assumptions in the object tracking algorithm with evidence-based logic, in detail Subjective Logic (SL) \cite{josang2016}. They calculated a degree of conflict to self-assess their tracking performance.

However, to the best of our knowledge, the information gained by fusing sensor data with evidence-based logic was not propagated to other stages of automated driving, like motion planning, but used only in their respective stage.

There exist various works that try to inherently handle incomplete or erroneous environments in the topic of motion planning like fail-safe approaches proposed in~\cite{Pek2018ComputationallyOptimization} or approaches using responsible safe sets as in~\cite{Shalev-Shwartz2017OnCars}.  
However, they use only an abstract version of the environment and do not consider the insights gained in previous stages.

Consequently, we aim to use the full potential of evidential grid maps in the subsequent motion planning stage.

\section{Foundations}
\label{sec:foundations}
This section briefly introduces the background and notation required for the derivation of our proposed method.
\subsection{Occupancy Grid Mapping}
\label{sec:foundations:grid_mapping}
\begin{figure}[tb]
\centering
\vspace{2mm}
\begin{minipage}[t]{0.47\columnwidth}
\centering
\begin{tikzpicture}
\begin{axis}[axis line style={draw=none}, tick style={draw=none}, yticklabels={,,}, xticklabels={,,}, width=3cm, height=1.7cm, legend columns=3]
\addplot[color=red, forget plot]{exp(x)};
\addlegendimage{only marks, mark size=0.1cm, mark=square*,color=black, fill=white}
\addlegendentry{\small{free}}
\addlegendimage{only marks, mark size=0.1cm, mark=square*,color=gray}
\addlegendentry{\small{unknown}}
\addlegendimage{only marks, mark size=0.1cm, mark=square*,color=black}
\addlegendentry{\small{occupied}}
\end{axis}
\end{tikzpicture}%
\end{minipage}%
\begin{minipage}[t]{0.47\columnwidth}
\centering
\begin{tikzpicture}
\begin{axis}[axis line style={draw=none}, tick style={draw=none}, yticklabels={,,}, xticklabels={,,}, width=\columnwidth, height=1.7cm, legend columns=2]
\addplot[color=red, forget plot]{exp(x)};
\addlegendimage{only marks, mark size=0.1cm, mark=square*,color=black, fill=white}
\addlegendentry{\small{drivable}}
\addlegendimage{only marks, mark size=0.1cm, mark=square*,color=black}
\addlegendentry{\small{non-drivable}}
\end{axis}
\end{tikzpicture}%
\end{minipage}%

\subfloat[]{\label{fig:occupancy}\def\svgwidth{0.46\columnwidth}
    \graphicspath{{img/foundations/grid_mapping}}
\begingroup%
  \makeatletter%
  \providecommand\color[2][]{%
    \errmessage{(Inkscape) Color is used for the text in Inkscape, but the package 'color.sty' is not loaded}%
    \renewcommand\color[2][]{}%
  }%
  \providecommand\transparent[1]{%
    \errmessage{(Inkscape) Transparency is used (non-zero) for the text in Inkscape, but the package 'transparent.sty' is not loaded}%
    \renewcommand\transparent[1]{}%
  }%
  \providecommand\rotatebox[2]{#2}%
  \newcommand*\fsize{\dimexpr\f@size pt\relax}%
  \newcommand*\lineheight[1]{\fontsize{\fsize}{#1\fsize}\selectfont}%
  \ifx\svgwidth\undefined%
    \setlength{\unitlength}{425.94712145bp}%
    \ifx\svgscale\undefined%
      \relax%
    \else%
      \setlength{\unitlength}{\unitlength * \real{\svgscale}}%
    \fi%
  \else%
    \setlength{\unitlength}{\svgwidth}%
  \fi%
  \global\let\svgwidth\undefined%
  \global\let\svgscale\undefined%
  \makeatother%
  \begin{picture}(1,0.39131581)%
    \lineheight{1}%
    \setlength\tabcolsep{0pt}%
    \put(0,0){\includegraphics[width=\unitlength,page=1]{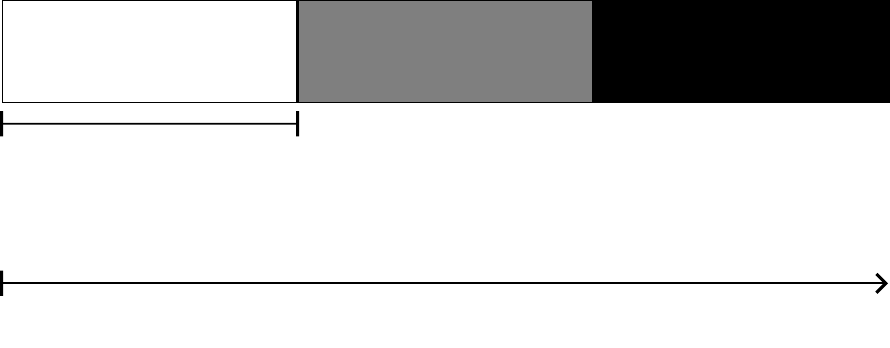}}%
    \put(0.150806,0.19999592){\color[rgb]{0,0,0}\makebox(0,0)[t]{\lineheight{1.25}\smash{\begin{tabular}[t]{c}$p_F$\end{tabular}}}}%
    \put(0,0){\includegraphics[width=\unitlength,page=2]{occupancy_classification.pdf}}%
    \put(0.30942119,0.1090875){\color[rgb]{0,0,0}\makebox(0,0)[t]{\lineheight{1.25}\smash{\begin{tabular}[t]{c}$p_U$\end{tabular}}}}%
    \put(0.50767628,0.00788833){\color[rgb]{0,0,0}\makebox(0,0)[t]{\lineheight{1.25}\smash{\begin{tabular}[t]{c}$p(x)$\end{tabular}}}}%
  \end{picture}%
\endgroup%
}%
 \subfloat[]{\label{fig:drivable}\def\svgwidth{0.46\columnwidth}
    \graphicspath{{img/foundations/grid_mapping}}
\begingroup%
  \makeatletter%
  \providecommand\color[2][]{%
    \errmessage{(Inkscape) Color is used for the text in Inkscape, but the package 'color.sty' is not loaded}%
    \renewcommand\color[2][]{}%
  }%
  \providecommand\transparent[1]{%
    \errmessage{(Inkscape) Transparency is used (non-zero) for the text in Inkscape, but the package 'transparent.sty' is not loaded}%
    \renewcommand\transparent[1]{}%
  }%
  \providecommand\rotatebox[2]{#2}%
  \newcommand*\fsize{\dimexpr\f@size pt\relax}%
  \newcommand*\lineheight[1]{\fontsize{\fsize}{#1\fsize}\selectfont}%
  \ifx\svgwidth\undefined%
    \setlength{\unitlength}{425.19689365bp}%
    \ifx\svgscale\undefined%
      \relax%
    \else%
      \setlength{\unitlength}{\unitlength * \real{\svgscale}}%
    \fi%
  \else%
    \setlength{\unitlength}{\svgwidth}%
  \fi%
  \global\let\svgwidth\undefined%
  \global\let\svgscale\undefined%
  \makeatother%
  \begin{picture}(1,0.39200631)%
    \lineheight{1}%
    \setlength\tabcolsep{0pt}%
    \put(0,0){\includegraphics[width=\unitlength,page=1]{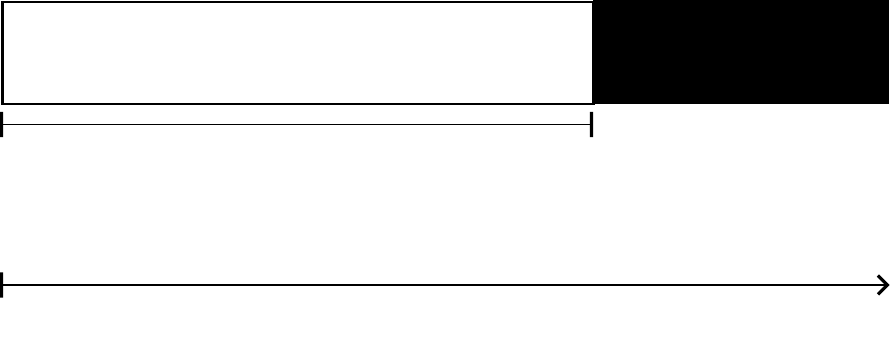}}%
    \put(0.31913139,0.20418824){\color[rgb]{0,0,0}\makebox(0,0)[t]{\lineheight{1.25}\smash{\begin{tabular}[t]{c}$p_D$\end{tabular}}}}%
    \put(0.50857209,0.00790225){\color[rgb]{0,0,0}\makebox(0,0)[t]{\lineheight{1.25}\smash{\begin{tabular}[t]{c}$p(x)$\end{tabular}}}}%
  \end{picture}%
\endgroup%
}
  \caption{%
  (a) Categorization into \textit{occupied}, \textit{unknown} and \textit{free} cells. 
  (b) Categorization into \textit{drivable} and \textit{non-drivable} cells.}
\end{figure}

Occupancy grid mapping (OGM) describes the process of creating a map $m=\{m_i\}$, which consists of many independent grid cells $m_i$ containing an estimate of their state of occupancy.
In conventional grid mapping approaches, Bayesian probabilities are used to estimate this state. 
In this case, $p(m_i)=0.0$ denotes a \textit{free} cell, $p(m_i)=1.0$ an \textit{occupied} cell, and $p(m_i)=0.5$ a cell with \textit{unknown} state, as shown in Fig.~\ref{fig:occupancy}\cite{thrun2005probabilistic}. 

For subsequent processing steps, the cells each containing a Bayesian probability are often categorized into \textit{free}, \textit{unknown} and \textit{occupied} by the thresholds $p_F$, $p_U$ as shown in Fig.~\ref{fig:occupancy} or into \textit{drivable} and \textit{non-drivable} by $p_D$ to be used for a path planning algorithm, as shown in Fig.~\ref{fig:drivable} \cite{thrun2005probabilistic}.

\subsection{Probabilistic Multi-Sensor Fusion}
\label{sec:foundations:probabilistic_multi_sensor_fusion}
Probabilistic multi-sensor fusion describes the process of combining multiple probabilistic measurements of cells originating from a number of sensors~$\mathbb{S}$.
\begin{definition}[Probabilistic Multi-Sensor Fusion \cite{thrun2005probabilistic}]
\label{def:multiSensor}
    Let $m^k=\{m^k_i\}$ be a grid map of sensor $k\in \mathbb{S}$ containing indexed cells $m_i^k$, where the probability of a cell $p(m_i^k)$ represents the probability of the cell being occupied.
    Now, the multi-sensor fusion can be defined in two ways.
    Measurements are either combined using the \textit{De Morgan's law} or by selecting the maximum probability.
    Formally, the multi-sensor fusion is defined by either
    \begin{subequations}
    \begin{alignat}{2}
    \label{eq:prod}
        p(m_i) & = 1 - \prod_{k \in \mathbb{S}} (1-p(m_i^k)) \quad \text{or} \\
    \label{eq:max}
        p(m_i) & = \max_k p(m_i^k)  \pspace ,
    \end{alignat}
    \end{subequations}
    respectively, where $p(m_i)$ denotes the fused cell information of all available sensor measurements.
\end{definition}
Both formulas in Definition~\ref{def:multiSensor} lead to conservative results.
Given that a sensor yields that a cell is occupied, Eq.~\eqref{eq:prod} causes every other measurement to further support this information, even when the other measurement would state a \textit{free} cell.
In contrast, Eq.~\eqref{eq:max} always selects the measurement predicting the highest probability of a cell being occupied.
Thus, once a cell is \textit{occupied}, measurements of a \textit{free} cell are simply ignored. 
Consequently, the case of conflicting sensor information caused by errors in the sensor setup cannot be modeled by this framework ~\cite{thrun2005probabilistic}.

In evidential grid mapping with DST or SL, uncertainty is modeled distinctively to allow a more elaborate approach for the fusion, as shown in the next section. 
SL explicitly models uncertainty, while with DST, uncertainty is implicitly represented by the lack of evidence.
Further, by considering uncertainty, the categorization can done more sophisticated, as shown in Sec.~\ref{sec:method:conflict-extraction}.

\subsection{Subjective Logic}
\label{sec:foundations:sl}
This section explains the fundamentals of SL used in this paper. 
For a more detailed explanation, the reader is referred to~\cite{josang2016}.
SL is a mathematical framework that can explicitly represent statistical uncertainty~\cite{josang2016}, comparable to the Dempster–Shafer theory~\cite{Shafer1976}.
In this work, we assume that input data based on SL is available, hence, mainly the representation of SL opinions and their interpretation is described in the following.
The fundamental aspect of SL is the representation of opinions.
A multinomial opinion incorporates information about a discrete random variable $X$ for every event $x$ in the sample space $\mathbb{X}$, regarding belief, uncertainty, and base rate.
\begin{definition}[Multinomial Opinion~\cite{josang2016}] \label{def:multinomial-opinion}
    Let $X$ be a random variable in the finite domain $\mathbb{X}$ with cardinality $|\mathbb{X}| \geq 2$. 
    A multinomial opinion is defined as an ordered triplet $\omega_X = (\vec{b}_X, u_X,\vec{a}_X)$ with
    \begin{subequations}
    	\begin{align}
    		\vec{b}_X(x) : \mathbb{X} \mapsto [0,1], \qquad 1 &= u_X + \sum\limits_{x \in \mathbb{X}} \vec{b}_X (x) \pspace , \\
    		\vec{a}_X(x) : \mathbb{X} \mapsto [0,1], \qquad 1 &= \sum\limits_{x \in \mathbb{X}} \vec{a}_X (x) \pspace .
    	\end{align}
    \end{subequations}
    The belief mass distribution $\vec{b}_X$ over $\mathbb{X}$ describes the belief in each event,
    while the uncertainty mass $u_X \in [0,1]$ models the lack of evidence.
    The base rate distribution $\vec{a}_X$ over $\mathbb{X}$ reflects the prior probability for each event.
\end{definition}
Further, the projected probability $\vec{P}_X(x): \mathbb{X} \mapsto [0,1]$ can be used to project a multinomial opinion into a probability distribution similar to Bayesian probabilities. 
It is defined by
\begin{align} \label{eqn:projected-probability}
    \vec{P}_X(x) = \vec{b}_X(x) + \vec{a}_X(x) \cdot u_X.
\end{align}

\begin{figure}[t]
    \centering
    \vspace{1mm}
	\includegraphics[width=0.8\columnwidth]{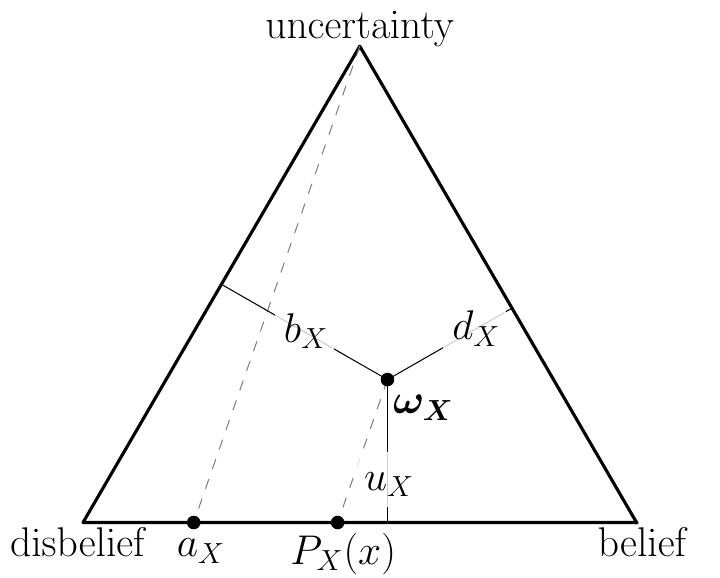}%
    \caption{%
        A binomial opinion $\omega_X$ is illustrated in a barycentric triangle. 
        The three axes of \textit{belief}, \textit{disbelief}, and \textit{uncertainty} are represented by $b_X$, $d_X$, and $u_X$, respectively, and $a_X$ is the prior projecting $\omega_X$ to $P_X(x)$.
    }
    \label{fig:bary}
\end{figure}

In this work, only the measurements of two LiDARs are fused. Hence, for the sake of simplicity in presentation, we focus on the description of binomial opinions, i.e., $\mathcal{X} = 2$ in the following.
They can be described based on the illustration in a barycentric triangle, which is depicted in Fig.~\ref{fig:bary}. Further, the additivity requirement $b_X + d_X + u_X = 1$ holds for every opinion $\omega_X = (b_X, d_X, u_X, a_X)$. 
However, all the presented methods are not limited to binomial opinions but are also possible with multinomial opinions.

%

In addition to that, opinions $\omega_X$ can be categorized not only by their projected probability $P_X(x)$, similar to Bayesian probabilities, but also depending on their uncertainty $u_X$. 
This leads to a two-dimensional categorization in the case of binomial opinions. 
The categorization itself is elaborated in~\cite{josang2016} and will be further addressed in Sec.~\ref{sec:method}.

\begin{figure}[tb]
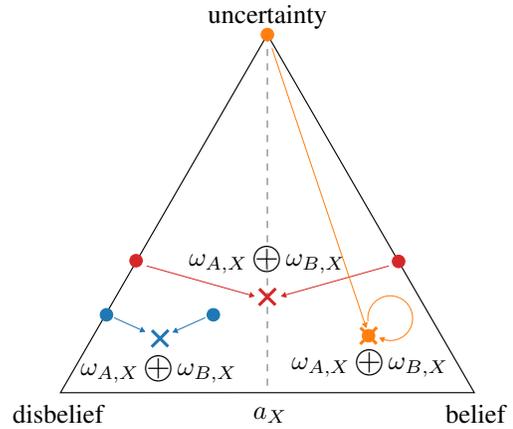

    \centering
    \vspace{3mm}
		\def\svgwidth{0.7\linewidth}
	\graphicspath{{img/foundations/sl}}
\begingroup%
  \makeatletter%
  \providecommand\color[2][]{%
    \errmessage{(Inkscape) Color is used for the text in Inkscape, but the package 'color.sty' is not loaded}%
    \renewcommand\color[2][]{}%
  }%
  \providecommand\transparent[1]{%
    \errmessage{(Inkscape) Transparency is used (non-zero) for the text in Inkscape, but the package 'transparent.sty' is not loaded}%
    \renewcommand\transparent[1]{}%
  }%
  \providecommand\rotatebox[2]{#2}%
  \newcommand*\fsize{\dimexpr\f@size pt\relax}%
  \newcommand*\lineheight[1]{\fontsize{\fsize}{#1\fsize}\selectfont}%
  \ifx\svgwidth\undefined%
    \setlength{\unitlength}{285.78486645bp}%
    \ifx\svgscale\undefined%
      \relax%
    \else%
      \setlength{\unitlength}{\unitlength * \real{\svgscale}}%
    \fi%
  \else%
    \setlength{\unitlength}{\svgwidth}%
  \fi%
  \global\let\svgwidth\undefined%
  \global\let\svgscale\undefined%
  \makeatother%
  \begin{picture}(1,0.90101168)%
    \lineheight{1}%
    \setlength\tabcolsep{0pt}%
    \put(0,0){\includegraphics[width=\unitlength,page=1]{fused_opinions.pdf}}%
    \put(0.27137174,0.10427735){\makebox(0,0)[ct]{\lineheight{1.25}\smash{\begin{tabular}[t]{l}$\omega_{A,X}\bigoplus\omega_{B,X}$\end{tabular}}}}%
    \put(0.5097865,0.34585859){\makebox(0,0)[ct]{\lineheight{1.25}\smash{\begin{tabular}[t]{l}$\omega_{A,X}\bigoplus\omega_{B,X}$\end{tabular}}}}%
    \put(0.73544582,0.11825542){\makebox(0,0)[ct]{\lineheight{1.25}\smash{\begin{tabular}[t]{l}$\omega_{A,X}\bigoplus\omega_{B,X}$\end{tabular}}}}%
    \put(0.51617222,0.00779815){\makebox(0,0)[ct]{\lineheight{1.25}\smash{\begin{tabular}[t]{l}$a_X$\end{tabular}}}}%
    \put(0.0497577,0.00064381){\makebox(0,0)[ct]{\lineheight{1.25}\smash{\begin{tabular}[t]{l}disbelief\end{tabular}}}}%
    \put(0.967794,0.00064381){\makebox(0,0)[ct]{\lineheight{1.25}\smash{\begin{tabular}[t]{l}belief\end{tabular}}}}%
    \put(0.51054566,0.88060466){\makebox(0,0)[ct]{\lineheight{1.25}\smash{\begin{tabular}[t]{l}uncertainty\end{tabular}}}}%
    \put(0,0){\includegraphics[width=\unitlength,page=2]{fused_opinions.pdf}}%
  \end{picture}%
\endgroup%

    \caption{
        The cumulative fusion of different opinions (dots) with their result (crosses) is illustrated in a barycentric triangle.
        The red dots are conflicting opinions; the blue dots are supporting ones.
        The fusion of an opinion with a neutral opinion with $u_X=1.0$ is shown in orange.
        }
    \label{fig:cumulative_fusion}
\end{figure}

In multi-sensor fusion based on SL opinions, the cumulative fusion operator $\bigoplus$ is often used.
It accumulates evidence in the evidence space and can handle conflicting information~\cite{josang2016}. 
Due to the accumulation, the uncertainty is reduced with every fusion of opinions, even when fused opinions are in conflict.
As a reference, the fusion of two conflicting opinions, of two supporting opinions, and of an one with a neutral opinion, are visualized in Fig.~\ref{fig:cumulative_fusion}.
\FloatBarrier
\section{Problem Formulation}\label{sec:problem_formulation}
AVs are equipped with several sensors to cover all areas around the vehicle. 
In some areas, the fields of view (FOV) of these sensors overlap to increase the robustness of the sensor setup in case of errors or occlusion of a sensor \cite{handbook}. 
\begin{figure}[b]
\centering
\scalebox{1.0}{
\begin{tikzpicture}
    \node[anchor=south west,inner sep=0] at (0,0) {\includegraphics[trim=17cm 9.0cm 17cm 8.5cm, clip, width=0.95\linewidth]{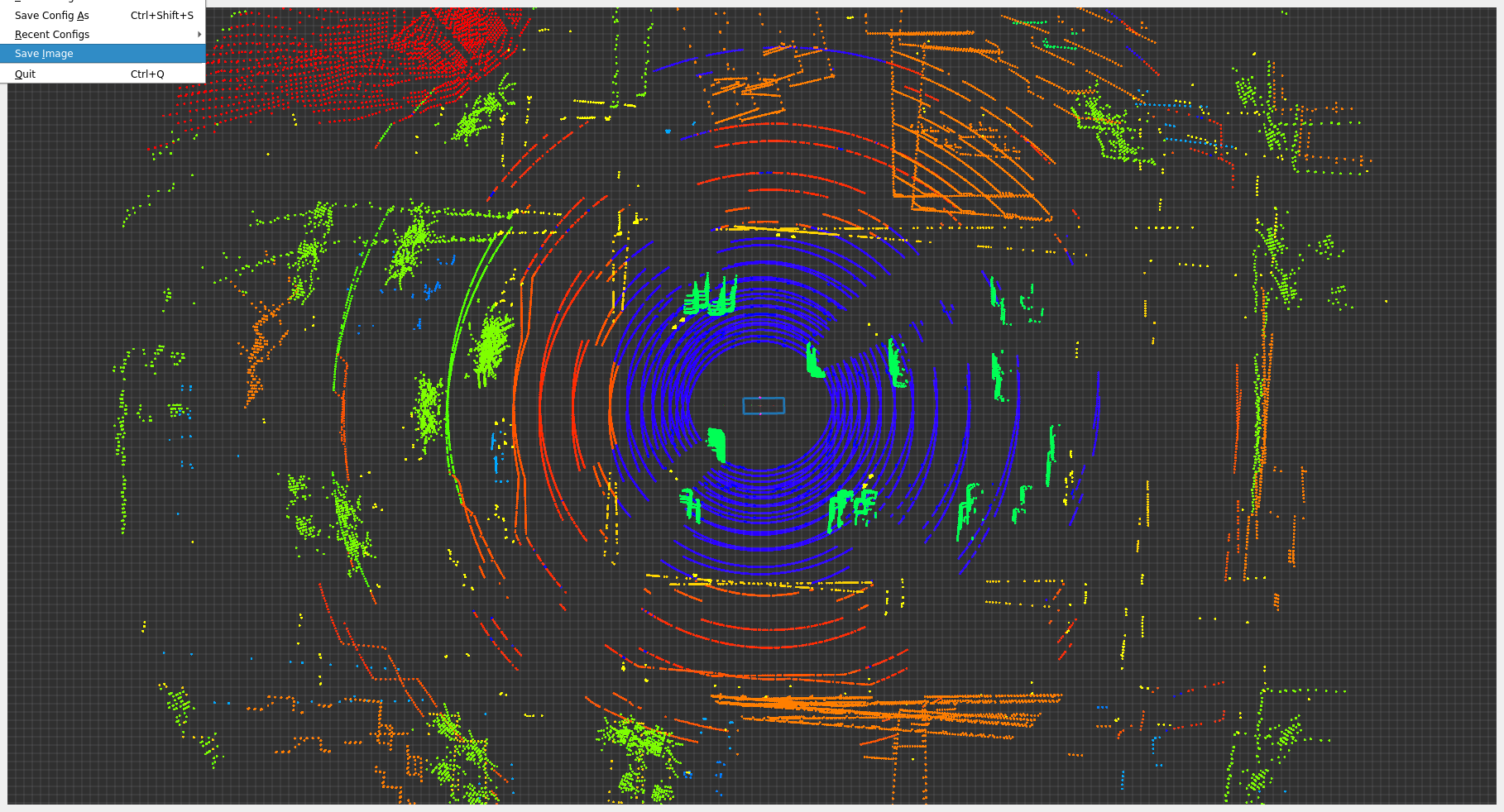}};
    \draw[TAB_RED,ultra thick,rounded corners] (2.4,3.5) rectangle (4.3,4.9);
\end{tikzpicture}}
\caption{
Point clouds of two LiDAR sensors mounted with a distance of \qty{1}{\metre} on top of a vehicle in the Carla simulator at a parking lot in \textit{Town04}.
The colors encode different object categories. 
One of the sensors has a rotational calibration error of \SI{5}{\degree} around its z-axis. 
The effect of the calibration error can, in particular, be observed on the parked cars in green and the street light poles in yellow (marked by the red rectangle).
}
\label{fig:point_cloud}
\end{figure}%

In this work, we investigate the impact on the overall system if sensors generate conflicting data.
Fig.~\ref{fig:point_cloud} shows an exemplary error of one sensor.
Here, the point clouds of two LiDAR sensors mounted onto a vehicle are visualized. 

However, one of the LiDAR sensors has a rotational calibration error of \SI{5}{\degree} around its z-axis.
It can be observed that this leads to the measurement of the same obstacles at different places. 
However, given no further information, it is impossible to determine which sensor data is correctly representing the reality.

\begin{figure}[t]
\centering
\vspace{6.5mm}
\raisebox{-0.2cm}[0pt][0pt]{%
\hspace{0.1mm}\begin{minipage}[t]{0.40\columnwidth}
\centering
\begin{tikzpicture}
\begin{axis}[axis line style={draw=none}, tick style={draw=none}, yticklabels={,,}, xticklabels={,,}, width=\textwidth, height=1.7cm, legend columns=2]
\addplot[color=red, forget plot]{exp(x)};
\addlegendimage{only marks, color=TAB_ORANGE}
\addlegendentry{LiDAR 1}
\addlegendimage{only marks, color=TAB_BLUE}
\addlegendentry{LiDAR 2}
\end{axis}
\end{tikzpicture}%
\end{minipage}%
\hspace{0.95em}
\begin{minipage}[t]{0.55\columnwidth}
\centering
\hspace{0.5mm}\begin{tikzpicture}
\begin{axis}[axis line style={draw=none}, tick style={draw=none}, yticklabels={,,}, xticklabels={,,}, width=3cm, height=1.7cm, legend columns=3]
\addplot[color=red, forget plot]{exp(x)};
\addlegendimage{only marks, mark size=0.15cm, mark=square*,color=black}
\addlegendentry{occupied}
\addlegendimage{only marks, mark size=0.15cm, mark=square*,color=gray}
\addlegendentry{unknown}
\addlegendimage{only marks, mark size=0.15cm, mark=square*,color=black, fill=white}
\addlegendentry{free}
\end{axis}
\end{tikzpicture}%
\end{minipage}%
}

\subfloat[]{\label{fig:points_overlay}\frame{\begin{tikzpicture}
    \node[anchor=south west,inner sep=0] at (0,0) {\includegraphics[trim=8cm 2.0cm 8cm 5.2cm, clip, width=0.48\columnwidth]{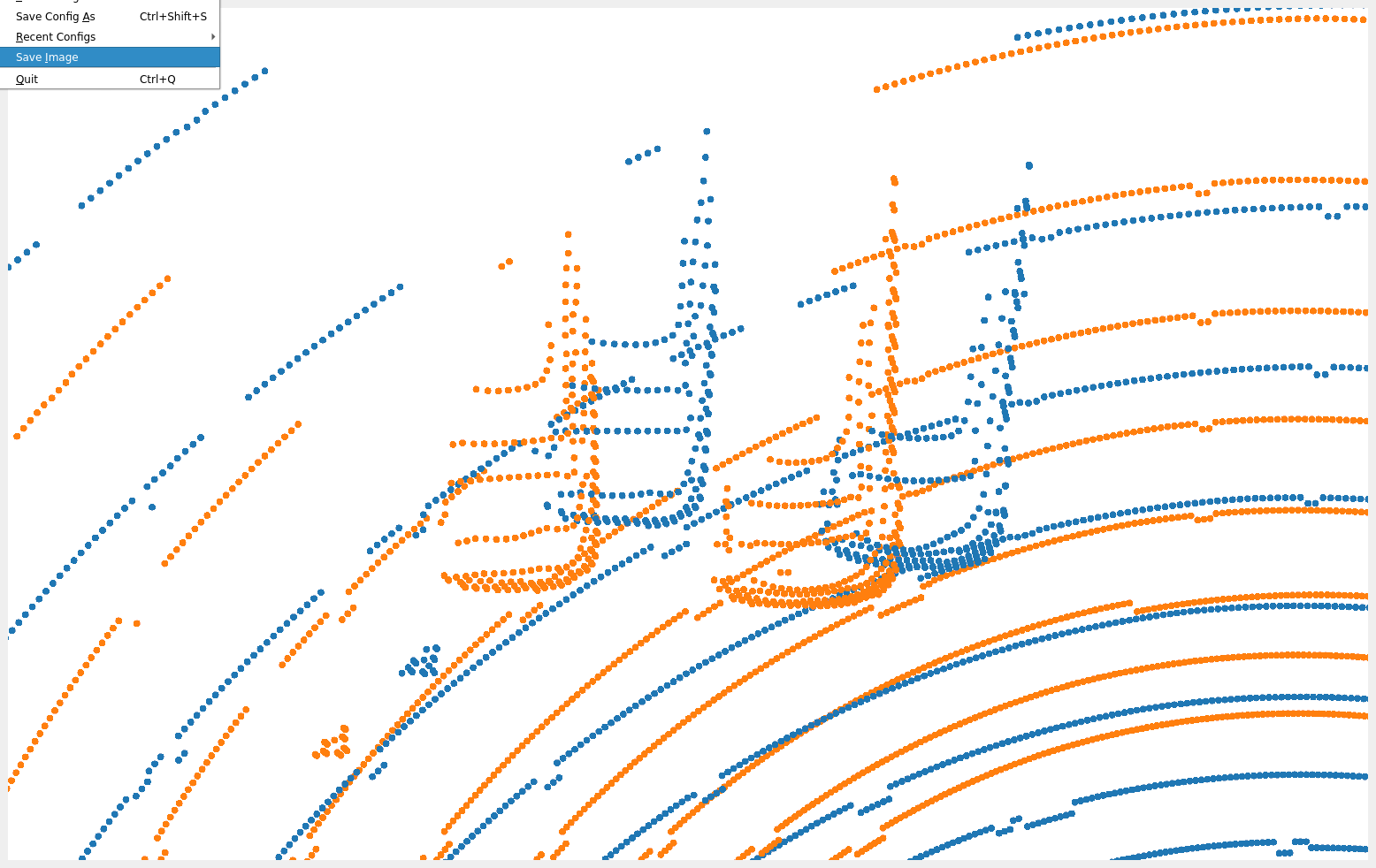}};
\draw[TAB_RED,ultra thick,rounded corners] (0.1,0.0) rectangle (1.0,1.0);
\draw[TAB_RED,ultra thick,rounded corners] (3.3,1.0) rectangle (4.0,3.0);
\end{tikzpicture}}}
  \hspace{0.5em}
 \subfloat[]{\label{fig:fused_grid}\frame{\begin{tikzpicture}
    \node[anchor=south west,inner sep=0] at (0,0) {\includegraphics[trim=9cm 6.7cm 8cm 12cm, clip, width=0.48\columnwidth]{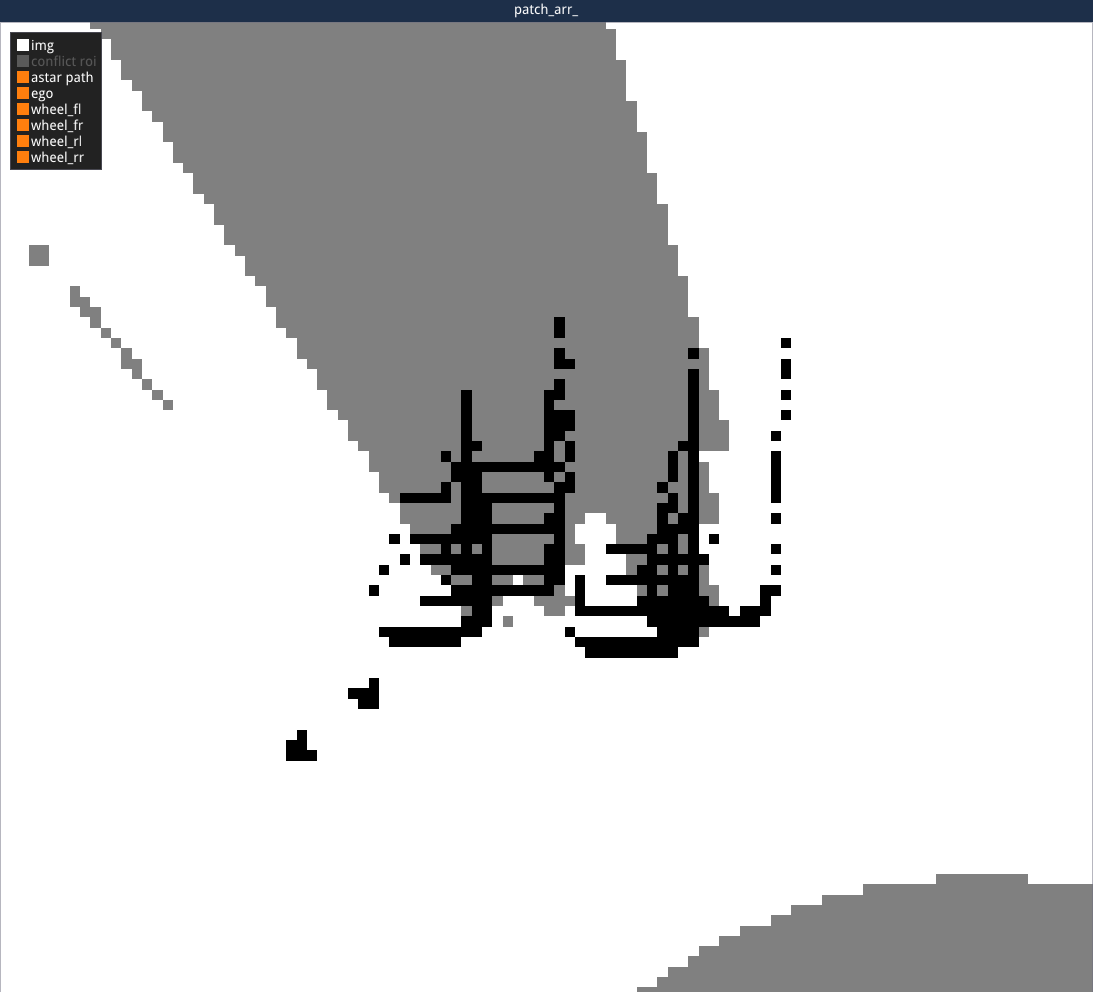}};
\draw[TAB_RED,ultra thick,rounded corners] (0.1,0.0) rectangle (1.0,1.0);
\draw[TAB_RED,ultra thick,rounded corners] (3.2,1.0) rectangle (4.0,3.0);
\end{tikzpicture}}}
  \caption{Exemplary scene with simulated measurements from two LiDAR sensors.
  (a) Point clouds of two different LiDAR sensors shown in blue and orange.
  The blue points are generated by the LiDAR which is not calibrated correctly. 
  (b) Due to this error, cells are measured to be \textit{free} by one sensor and \textit{occupied} by the other. 
  Due to the fusion using \textit{De Morgan's Law} of Bayesian probabilities, every cell that contains a colliding LiDAR measurement is classified as \textit{occupied}.}
\end{figure}
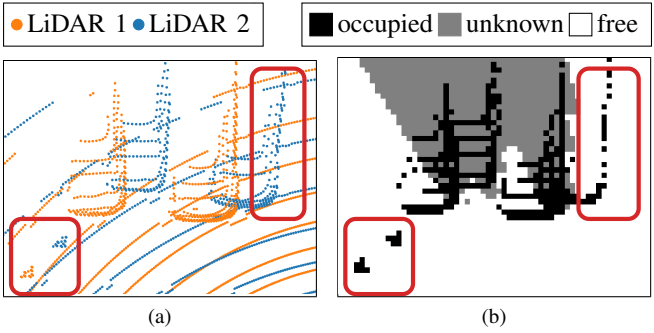

In conventional occupancy grid mapping, the measurements of two sensors are fused into one cell by the mentioned fusion of probabilities by \textit{De Morgan's law} as shown in Section~\ref{sec:foundations:grid_mapping}. 
This is shown in Fig.~\ref{fig:points_overlay} and \ref{fig:fused_grid}. 
Here, the conflicting measurements lead to a very conservative result of every cell containing conflicting measurements as being \textit{occupied}. Therefore, these cells will be classified as \textit{non-drivable}.

Another problem with the fusion of Bayesian probabilities is the loss of information. After the fusion, the information that the cell contained conflicting measurements is irretrievably lost.
However, we presume that it is critical to preserve this information of conflicting information in the perception stage and pass it through to the consecutive stages to be able to react properly.

\FloatBarrier
\section{Method}\label{sec:method}
The following subsections elaborate on the main points of conflict-aware motion planning on evidential grid maps. They build on one another and are shown in Fig.~\ref{fig:overview}.
\subsection{Conflict-Aware Classification of Grid Cells}\label{sec:method:conflict-extraction}
First, the conflict-aware classification of cells in an evidential grid map is proposed.
In this work, the adaptive patched grid map from \cite{wodtko2023adaptive} is used, in which the state of each cell is modeled by a binomial opinion $\omega(m_i)$ as stated in Definition~\ref{def:multinomial-opinion}.
Here, the measurements of different sensors are fused by cumulative fusion as shown in Fig.~\ref{fig:cumulative_fusion}. 
Hence, in this work, the already fused data is used.

\begin{figure}[b]
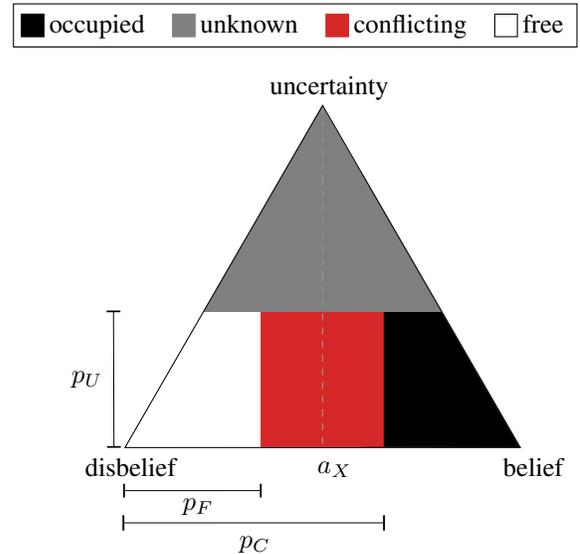

\vspace{2mm}
	\centering
    \begin{tikzpicture}
    \begin{axis}[axis line style={draw=none}, tick style={draw=none}, yticklabels={,,}, xticklabels={,,}, width=3cm, height=1.7cm, legend columns=4,
    legend style={/tikz/every even column/.append style={column sep=0.3cm}}
    ]
    \addplot[color=red, forget plot]{exp(x)};
    \addlegendimage{only marks, mark size=0.15cm, mark=square*,color=black}
    \addlegendentry{occupied}
    \addlegendimage{only marks, mark size=0.15cm, mark=square*,color=gray}
    \addlegendentry{unknown}
    \addlegendimage{only marks, mark size=0.15cm, mark=square*,color=TAB_RED}
    \addlegendentry{conflicting}
    \addlegendimage{only marks, mark size=0.15cm, mark=square*,color=black, fill=white}
    \addlegendentry{free}
    \end{axis}
    \end{tikzpicture}
    \vspace{0.5cm}

	\def\svgwidth{0.8\linewidth}
	\graphicspath{{img/method/categories}}
\begingroup%
  \makeatletter%
  \providecommand\color[2][]{%
    \errmessage{(Inkscape) Color is used for the text in Inkscape, but the package 'color.sty' is not loaded}%
    \renewcommand\color[2][]{}%
  }%
  \providecommand\transparent[1]{%
    \errmessage{(Inkscape) Transparency is used (non-zero) for the text in Inkscape, but the package 'transparent.sty' is not loaded}%
    \renewcommand\transparent[1]{}%
  }%
  \providecommand\rotatebox[2]{#2}%
  \newcommand*\fsize{\dimexpr\f@size pt\relax}%
  \newcommand*\lineheight[1]{\fontsize{\fsize}{#1\fsize}\selectfont}%
  \ifx\svgwidth\undefined%
    \setlength{\unitlength}{341.89264973bp}%
    \ifx\svgscale\undefined%
      \relax%
    \else%
      \setlength{\unitlength}{\unitlength * \real{\svgscale}}%
    \fi%
  \else%
    \setlength{\unitlength}{\svgwidth}%
  \fi%
  \global\let\svgwidth\undefined%
  \global\let\svgscale\undefined%
  \makeatother%
  \begin{picture}(1,0.89958044)%
    \lineheight{1}%
    \setlength\tabcolsep{0pt}%
    \put(0,0){\includegraphics[width=\unitlength,page=1]{categories.pdf}}%
    \put(0.53196553,0.152819){\makebox(0,0)[lt]{\lineheight{1.25}\smash{\begin{tabular}[t]{l}$a_X$\end{tabular}}}}%
    \put(0.0848481,0.14683876){\makebox(0,0)[lt]{\lineheight{1.25}\smash{\begin{tabular}[t]{l}disbelief\end{tabular}}}}%
    \put(0.88941891,0.14683876){\makebox(0,0)[lt]{\lineheight{1.25}\smash{\begin{tabular}[t]{l}belief\end{tabular}}}}%
    \put(0.44023879,0.8825224){\makebox(0,0)[lt]{\lineheight{1.25}\smash{\begin{tabular}[t]{l}uncertainty\end{tabular}}}}%
    \put(0,0){\includegraphics[width=\unitlength,page=2]{categories.pdf}}%
    \put(0.30207635,0.08072978){\makebox(0,0)[t]{\lineheight{1.25}\smash{\begin{tabular}[t]{c}$p_F$\end{tabular}}}}%
    \put(0.41180743,0.00892682){\makebox(0,0)[t]{\lineheight{1.25}\smash{\begin{tabular}[t]{c}$p_C$\end{tabular}}}}%
    \put(0,0){\includegraphics[width=\unitlength,page=3]{categories.pdf}}%
    \put(0.05898637,0.326532){\makebox(0,0)[lt]{\lineheight{1.25}\smash{\begin{tabular}[t]{l}$p_U$\end{tabular}}}}%
  \end{picture}%
\endgroup%

\caption{Barycentric triangle with the binomial opinion of $belief$ and $uncertainty$. 
In contrast to conventional classification of grid cells, an additional class \textit{conflicting} is available, which is mapped to regions with a low uncertainty but conflicting beliefs.}
\label{fig:categories}
\end{figure}

Next, each binomial opinion is classified into one of four categories similar to the categories proposed in \cite{josang2016}. 
They are \textit{unknown}~(\textit{U}), \textit{free}~(\textit{F}), \textit{conflict}~(\textit{C}) and \textit{occupied}~(\textit{O}). 
This classification can be described by the mapping function
\begin{align}
    f(\omega(m_i)) = \begin{cases}
U &\text{for } u_X > p_U, \\
\multirow{2}{*}{F} &\text{for }u_X < p_U \\ 
                   &\text{and }P_X(m_i) \le p_F, \\
\multirow{2}{*}{C} &\text{for }u_X < p_U \\ 
                   &\text{and }p_F < P_X(m_i) < p_C, \\
\multirow{2}{*}{O} &\text{for }u_X < p_U \\ 
                   &\text{and }P_X(m_i) \ge p_C \pspace. \\
\end{cases}
\label{eq:categorization}
\end{align}
This two-dimensional classification of opinions is visualized in the barycentric triangle in Fig.~\ref{fig:categories}.
Here, the base rate is set to $a_X=0.5$.

\subsection{Self-Assessment}\label{sec:method:self-asessment}
Next, the categorized grid map is used to allow the general evaluation of the data generated by the sensor system.

A prerequisite for this evaluation is the dilation of the mentioned grid map similar to the dilation used in conventional collision checking as presented in \cite{Ziegler2010FastPlanning}. 
We do this to measure the impact of the \textit{conflicting} cells on path planning approaches that use dilated versions of their environment to allow the efficient evaluation of vehicle poses.
Therefore, the values decoding the mentioned categories are ordered from high values to low values $v\textsubscript{O} > v\textsubscript{C} > v\textsubscript{U} > v\textsubscript{F}$. 
This is necessary as during the dilation process, cells with high values dilate over cells with low values, meaning that occupied cells will be dilated over \textit{conflicting} cells, \textit{conflicting} ones over \textit{unknown} ones, and \textit{unknown} ones over \textit{free} ones. 
The dilation of an exemplary grid map is shown in Fig.~\ref{fig:dilated_scene}.

Using the dilated grid map, we propose the estimation of a so-called degradation score.
It is calculated using the number of \textit{conflicting} cells $n\textsubscript{c}$ and the number of \textit{occupied} cells $n\textsubscript{o}$ within a maximum distance $d\textsubscript{max}$ to the ego vehicle. This distance defines the region of interest (ROI) of the method.
These conflicting and occupied cells are weighted by a function of their distance $g(d): d \mapsto [0,1]$ to the ego vehicle. 
The weighting function is given by 
\begin{align}
    g(d) = \begin{cases}
\frac{d\textsubscript{max} - d}{d\textsubscript{max}}  & \text{for } d \leq d\textsubscript{max} \\
0 & \text{for } d > d\textsubscript{max} \pspace.
\end{cases}
\end{align}
The closer cells are to the vehicle, the higher their weight, which leads to the calculation of the degradation score
\begin{align}
    \alpha = \frac{\sum_{i}^{n\textsubscript{c}} g(d_i)}{\sum_{i}^{n\textsubscript{c}} g(d_i) + \sum_{i}^{n\textsubscript{o}} g(d_i)} \pspace.
\end{align}
A value of 0.0 indicates that there are no conflicting cells in the current ROI, while a value of 1.0 means that all cells that would have been \textit{occupied} by conventional grid mapping are \textit{conflicting}. 
If there are neither \textit{occupied} nor \textit{conflicting} cells, $\alpha$ is unbound and cannot be interpreted as no conclusion about conflict can be drawn from this environment model.
If $\alpha$ is higher than a certain value $\alpha\textsubscript{max}$, the vehicle can derive the following actions from it:
\begin{enumerate} 
\item The system is no longer inside its operational design domain (ODD). Hence, the first step is to alert the driver, if present, or the responsible monitoring unit of the vehicle.
Further, depending on the mission requirements and safety regulations, the vehicle must execute
\item[2a)] an emergency stop, or it can
\item[2b)] proceed to fulfill its mission if possible or reach a safe pose while using the additional information.
\end{enumerate}
Further, each pose of a planned path can be evaluated on the dilated categorized grid map to specify if the pose is affected by the degraded environment model.
If $\alpha$ is high, but the planned path does not pass over conflicting cells, the motion plan of the robot is not affected by the degradation. This means it can continue to follow its path.

\subsection{Conflict-Aware Path Planning}
\label{sec:method:path-planning}
The proposed method of conflict-aware path planning is based on a derivation of the graph-based path planning algorithm presented in \cite{schumann2023efficient}. 
It is an adaption of the Hybrid~A*-algorithm~\cite{Dolgov2008PracticalDriving} that was augmented with an early-stopping strategy. 
This algorithm was chosen because graph-based path planning algorithms, as well as sampling-based algorithms \cite{Klemm2015}, have the advantage of being able to use non-differentiable costs in the form of a cost map during the planning process. 

When planning through areas with conflicting measurements, one must consider the following:
The binomial opinion of the cells can vary depending on the position of the observing sensor. 
For example, the impact of rotational calibration errors is higher the further away the sensor is. With decreasing distance, the spatial impact of this error decreases. 
Hence, if the robot is far away from \textit{conflicting} cells, there is a chance that the number of \textit{conflicting} cells will decrease while driving toward them. Hence, approaching a region with conflicting data is worth a try. 

However, poses in conflicting areas should be avoided if possible. This is done by using the dilated categorized grid map as a cost map. 
Path segments that overlap with this map are assigned additional movement cost of $c\textsubscript{C}$ per meter.
Further, these segments must be approached cautiously and conflicting poses itself should not be passed over at all. This can be implemented by velocity constraints inside the underlying trajectory planner.

Nevertheless, translational calibration errors generate a constant spatial error that does not decrease when approaching an obstacle.
Therefore, if the distance of \textit{conflicting} cells to the vehicle lies under a certain threshold $d\textsubscript{c}$, the cells must be considered \textit{occupied}. 
This will trigger a replanning of the path planning algorithm, which then avoids these areas completely.
To conclude, conflict-aware path planning consists of the following steps:
\begin{enumerate}
    \item The path planning algorithm is allowed to pass \textit{conflicting} cells;
    \item the movement cost of path segments that pass \textit{conflicting} cells is increased;
    \item path segments over \textit{conflicting} cells are marked for the underlying trajectory planning algorithm, which must not pass these segments; and
    \item \textit{conflicting} cells with $d_i < d\textsubscript{c}$ are marked as \textit{occupied}.
\end{enumerate}
This leads to a curious and robust path planning strategy that aims at preventing the unnecessary abortion of planning tasks.

\FloatBarrier
\section{Evaluation}\label{sec:eval}
For the evaluation of the proposed methods, the parameters from Table~\ref{tab:params} are used.
\begin{table}[b]
	\caption{Evaluation Setup}
 \vspace{-0.5cm}
	\label{tab:params}
	\begin{center}
		\begin{tabular}{c | l | c}
			\toprule
			Parameter & Description & Value \\
			\midrule
   $a\textsubscript{X}$ & base rate of binomial opinions & $0.5$ \\
    $p\textsubscript{U}$ & threshold for \textit{unknown} cells & $0.3$ \\
    $p\textsubscript{F}$ & threshold for \textit{free} cells & $0.2$ \\
    $p\textsubscript{C}$ & threshold for \textit{conflicting} cells & $0.8$ \\
   $d\textsubscript{max}$ & max. distance to evaluate \textit{conflicting} cells & \qty{15}{\meter} \\
  $d\textsubscript{c}$ & conflict. with $d<d\textsubscript{c}$ are set to \textit{occupied} & \qty{5}{\meter} \\
  $c\textsubscript{C}$ & add. cost for poses in \textit{conflict.} cells per \qty{}{\metre} & 5.0 \\
      $n\textsubscript{ch}$  & number of channels of each LiDAR & 32 \\
    $PPS_0$  & points per second of standard LiDAR & 1310720 \\
    $PPS_1$  & points per second of insufficient LiDAR & 327680 \\
			\bottomrule
		\end{tabular}
	\end{center}
\end{table}%

\subsection{Conflict-Aware Classification of Grid Cells}
\begin{figure}[t]
\centering
\vspace{7mm}
 \raisebox{-0.2cm}[0pt][0pt]{%
\hspace{0.5mm}\begin{tikzpicture}
\begin{axis}[axis line style={draw=none}, tick style={draw=none}, yticklabels={,,}, xticklabels={,,}, width=1.17\columnwidth, height=1.6cm, legend columns=4,
legend style={/tikz/every even column/.append style={column sep=0.3cm}}
]
\addplot[color=red, opacity=0.0, forget plot]{x};
\addlegendimage{only marks, mark size=0.15cm, mark=square*,color=black}
\addlegendentry{occupied}
\addlegendimage{only marks, mark size=0.15cm, mark=square*,color=gray}
\addlegendentry{unknown}
\addlegendimage{only marks, mark size=0.15cm, mark=square*,color=TAB_RED}
\addlegendentry{conflicting}
\addlegendimage{only marks, mark size=0.15cm, mark=square*,color=black, fill=white}
\addlegendentry{free}
\end{axis}
\end{tikzpicture}}
  \subfloat[]{\label{fig:colored_conflict}\frame{\includegraphics[trim=15.5cm 20.2cm 19.5cm 13cm, clip, width=0.48\linewidth]{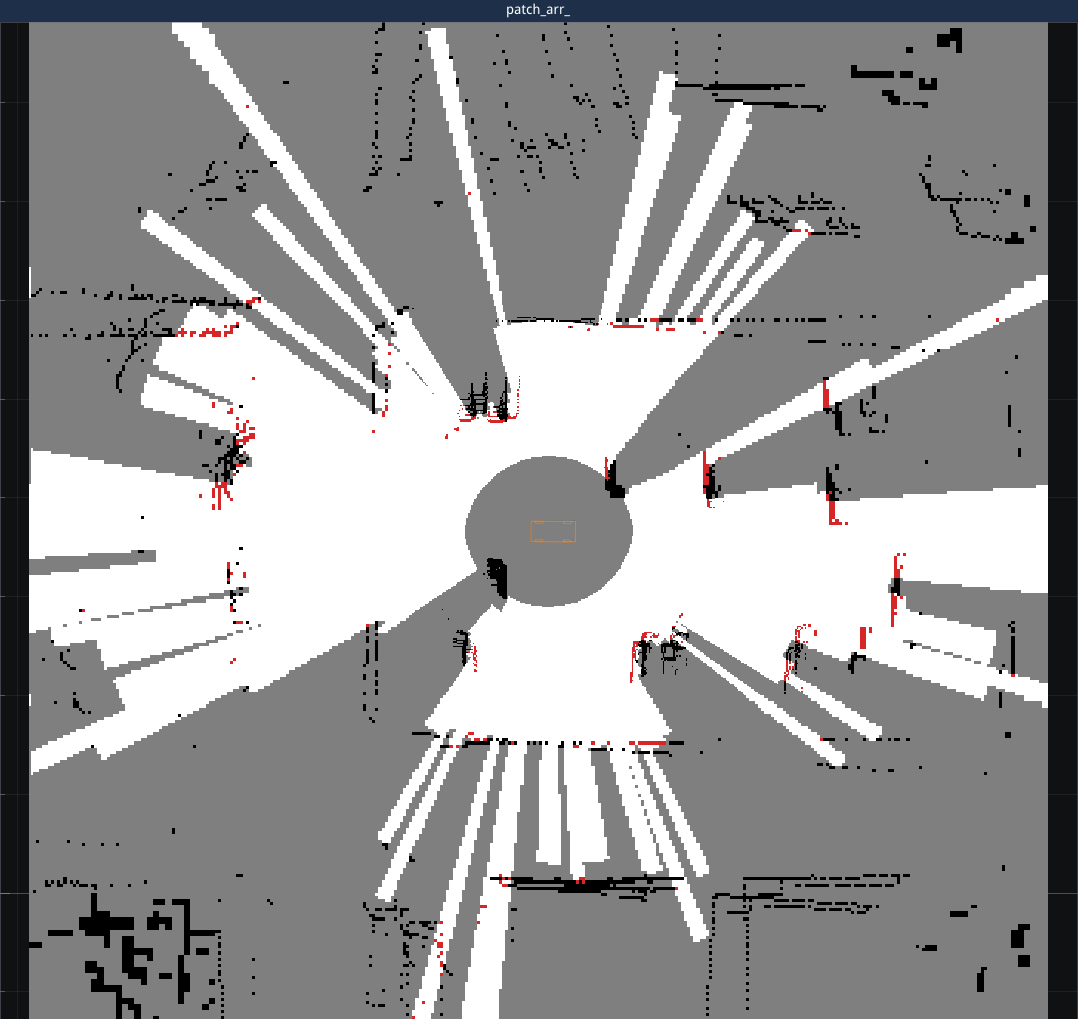}}}
  \hspace{0.5em}
 \subfloat[]{\label{fig:dilated_scene}\frame{\includegraphics[trim=6cm 7.5cm 7cm 10cm, clip, width=0.48\linewidth]{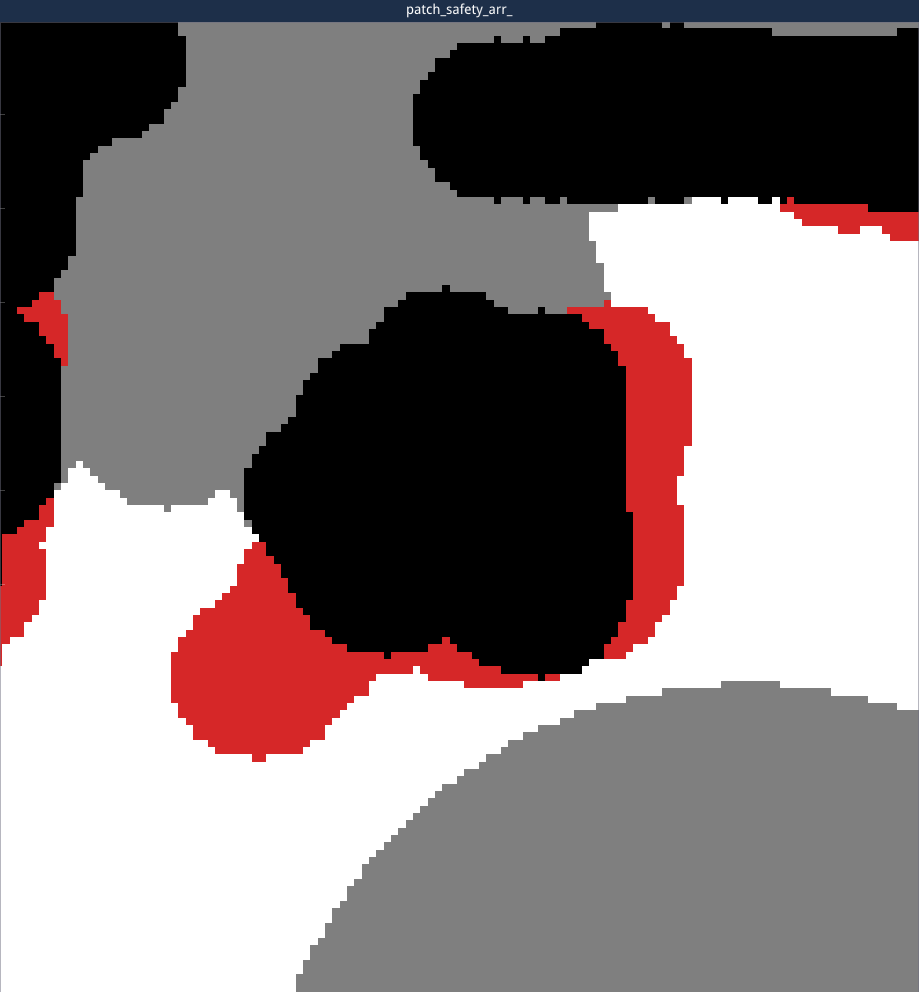}}}
  \caption{Conflict-aware classification of cells for the measurement shown in Fig.~\ref{fig:points_overlay}. (a) Cells containing \textit{conflicting} opinions are classified as \textit{conflicting}. 
  (b) The dilated map on which the degradation score $\alpha$ is calculated, and poses during the path planning process are evaluated.}
\end{figure}
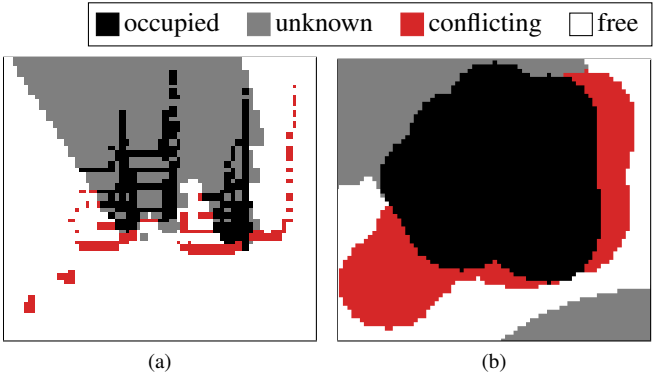

First, the conflict-aware classification is qualitatively evaluated in Fig.~\ref{fig:colored_conflict}, which shows the same scene as Fig.~\ref{fig:fused_grid}.
The results show that the cells representing the street light post and the edge of the parked cars are now classified as \textit{conflicting}. 
The results of the other cells are equal to the conventional method, leading to \textit{occupied} cells where both LiDARs measure an object and \textit{unknown} cells behind the parked cars. 
Fig.~\ref{fig:dilated_scene} shows the dilated map, which is used in the following section.

\subsection{Self-Assessment}
\begin{figure}[b]
\centering
\begin{tikzpicture}
    \pgfplotsset{
        legend pos=north west,
        width=1.0\columnwidth,
        height=6cm,
        ymin=0, ymax=0.4,
        x axis style/.style={
            xticklabel style=#1,
            xlabel style=#1,
            x axis line style=#1,
            xtick style=#1
       }
    }
    
    \begin{axis}[
      axis x line*=bottom,
      x axis style=TAB_BLUE,
      xmin=0, xmax=15,
      xlabel=rotational error in \SI{}{\degree},
      ylabel=degradation score $\alpha$,
      ymajorgrids=true,
      grid style=dashed,
    ]

    \addplot[forget plot, line width=1pt,solid,color=TAB_BLUE] %
    	table[x=yawerror, y=yaw_score_1,col sep=comma]{graphs/data/conflict_scores.csv};
     \addplot[forget plot, line width=1pt,solid,color=TAB_BLUE] %
    	table[x=yawerror, y=yaw_score_2,col sep=comma]{graphs/data/conflict_scores.csv};
     \addplot[forget plot, line width=1pt,solid,color=TAB_BLUE] %
    	table[x=yawerror, y=yaw_score_3,col sep=comma]{graphs/data/conflict_scores.csv};
      \addplot[forget plot, line width=1pt,dashed,color=TAB_BLUE] %
        table[x=yawerror, y=yaw_score_4,col sep=comma]{graphs/data/conflict_scores.csv};
      \addplot[forget plot, line width=1pt,dashed,color=TAB_BLUE] %
        table[x=yawerror, y=yaw_score_5,col sep=comma]{graphs/data/conflict_scores.csv};

    \addlegendentry{urban}
    \addlegendimage{solid, black}
    \addlegendentry{highway}
    \addlegendimage{dashed, black}

    \end{axis}
    
    \begin{axis}[
      axis x line*=top,
      axis y line=none,
      xmin=0, xmax=2.34375,
      xlabel=translational error in \SI{}{\metre},
      x axis style=TAB_RED
    ]

\addplot[line width=1pt,solid,color=TAB_RED] %
	table[x=xoffset, y=offset_score_1
,col sep=comma]{graphs/data/conflict_scores.csv};
 \addplot[line width=1pt,solid,color=TAB_RED] %
	table[x=xoffset, y=offset_score_2
,col sep=comma]{graphs/data/conflict_scores.csv};
 \addplot[line width=1pt,solid,color=TAB_RED] %
	table[x=xoffset, y=offset_score_3
,col sep=comma]{graphs/data/conflict_scores.csv};
 \addplot[line width=1pt,dashed,color=TAB_RED] %
	table[x=xoffset, y=offset_score_4
,col sep=comma]{graphs/data/conflict_scores.csv};
 \addplot[line width=1pt,dashed,color=TAB_RED] %
	table[x=xoffset, y=offset_score_5
,col sep=comma]{graphs/data/conflict_scores.csv};
\end{axis}
\end{tikzpicture}
\caption{With rising rotational (blue) and translational (red) calibration errors, the degradation score increases in all environments. The scores deviate depending on the environment. 
The scores are calculated in five environments, of which three are urban and two on a highway.}
\label{fig:conflict_scores}
\end{figure}
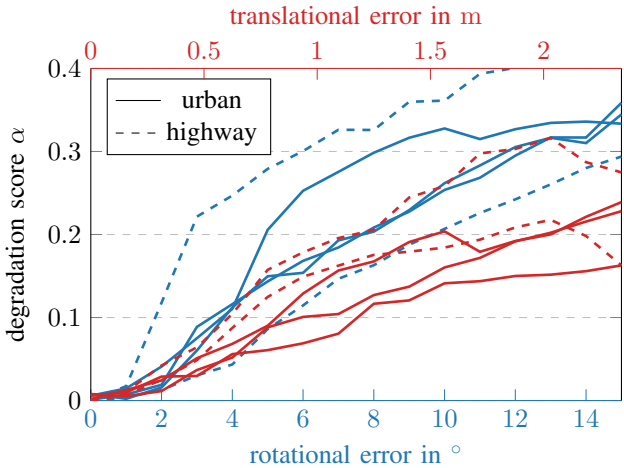

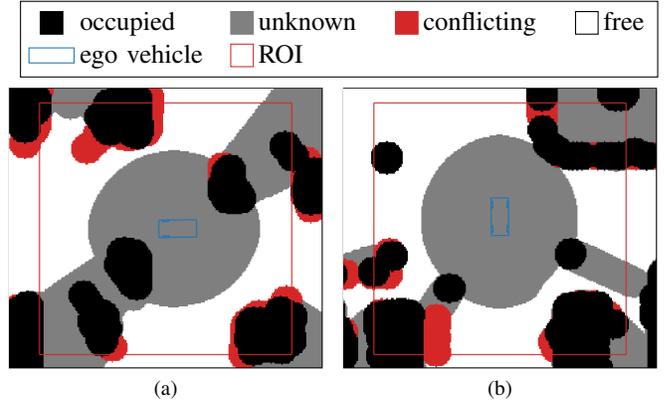
\begin{figure}[tb]
\centering
\vspace{11mm}
\raisebox{-0.2cm}[0pt][0pt]{%
\hspace{0.5mm}\begin{tikzpicture}
\begin{axis}[axis line style={draw=none}, tick style={draw=none}, yticklabels={,,}, xticklabels={,,}, width=1.17\columnwidth, height=1.6cm, legend columns=4,
legend cell align={left},
legend style={/tikz/every even column/.append style={column sep=0.3cm}}
]
\addplot[color=red, opacity=0.0, forget plot]{x};
\addlegendimage{only marks, mark size=0.15cm, mark=square*,color=black}
\addlegendentry{occupied}
\addlegendimage{only marks, mark size=0.15cm, mark=square*,color=gray}
\addlegendentry{unknown}
\addlegendimage{only marks, mark size=0.15cm, mark=square*,color=TAB_RED}
\addlegendentry{conflicting}
\addlegendimage{only marks, mark size=0.15cm, mark=square*,color=black, fill=white}
\addlegendentry{free}
\addlegendimage{area legend, color=TAB_BLUE}
\addlegendentry{ego vehicle}
\addlegendimage{only marks, mark size=0.15cm, mark=square*,color=TAB_RED, fill=white}
\addlegendentry{ROI}
\addlegendimage{empty legend}
\addlegendentry{}
\addlegendimage{empty legend}
\addlegendentry{}
\end{axis}
\end{tikzpicture}}
  \subfloat[]{\label{fig:conflict_score_calc}\frame{\includegraphics[trim=6cm 8cm 6cm 9.5cm, clip, width=0.48\linewidth]{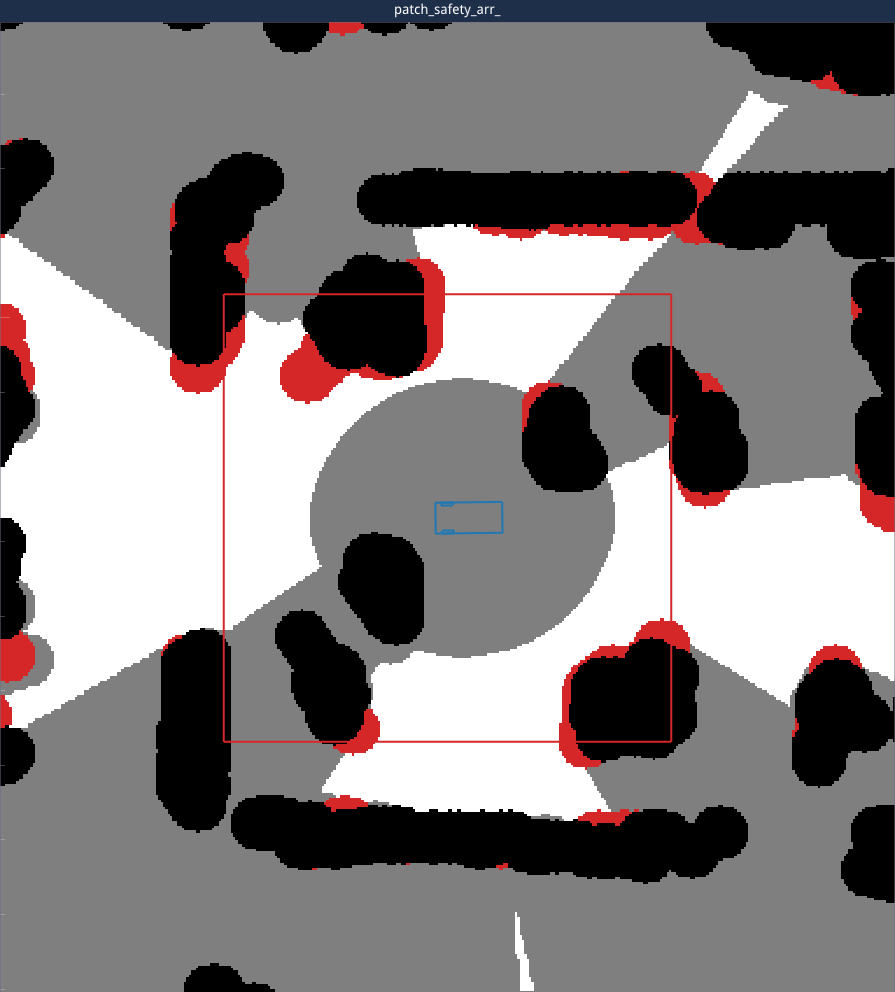}}}
  \hspace{0.5em}
 \subfloat[]{\label{fig:intrinsic_conflict}\frame{\includegraphics[trim=6cm 8cm 6cm 9.5cm, clip, width=0.48\linewidth]{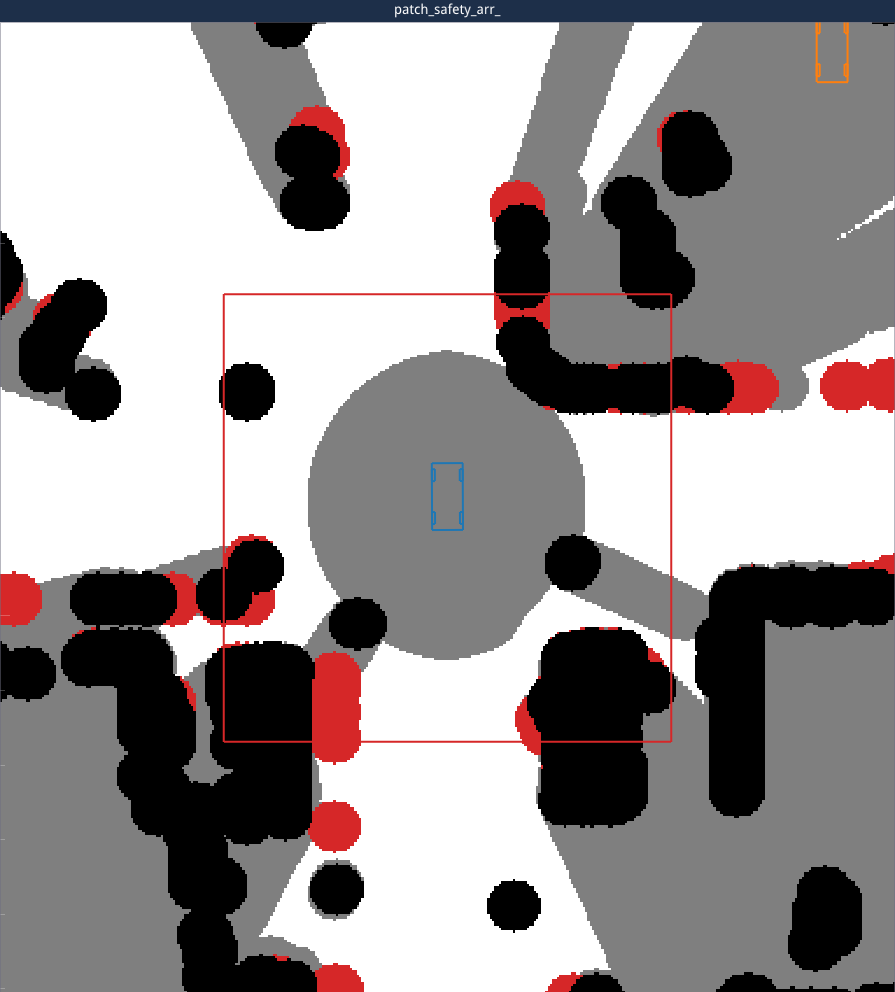}}}
  \caption{
  Dilated categorized evidential grid maps caused by different reasons. The gray area around the vehicle is caused by the LiDAR sensors that cannot reach the floor close to the vehicle.
  Hence, no cell can be measured as being free.
  (a) A degradation score of $\alpha \approx 0.09$ is calculated on the full scene of Fig.~\ref{fig:dilated_scene} due to a rotational calibration error of \SI{5}{\degree}. \\
  (b) A degradation score of $\alpha \approx 0.07$ calculated on another scene. Due to the insufficient angular resolution of the LiDARs, they are unable to detect the fences at the lower left and upper right reliably.
  }
\end{figure}

The degradation score $\alpha$ is now evaluated for rotational and translational calibration errors in five different environments in \textit{Town04} in the Carla simulator. Fig.~\ref{fig:conflict_scores} shows that $\alpha$ increases in all environments with rising errors. 
This shows that the self-assessment module can reliably detect erroneous sensor systems in various environments.
An exemplary environment generating a degradation score of $\alpha \approx 0.09$  is visualized in Fig.~\ref{fig:conflict_score_calc}.
In our experiments, we found that environment representations with a degradation score of $\alpha > 0.1$ are degraded enough to execute the steps presented in Sec.~\ref{sec:method:self-asessment} as in this case, every tenth measurement is causing \textit{conflicting} cells.

Further, some environments induce \textit{conflicting} cells inherently without errors in the sensor setup, which is shown in Fig.~\ref{fig:intrinsic_conflict}. 
Here, the angular resolution of the LiDARs was decreased by reducing the points per second (PPS) of the LiDARs from $PPS_0$ to $PPS_1$ by a factor of four as stated in Table~\ref{tab:params}.
Now, the thin poles of fences at the side of the road are not reliably detected by the LiDARs as some rays pass the gaps of the fence without colliding with them, leading to \textit{conflicting} measurements.
Thus, a degradation score of $\alpha \approx 0.07$ is calculated, which shows that even though the sensor setup is working as intended, it cannot measure the environment correctly. 

Therefore, we propose using this method to investigate sensor setups in their intended environment to validate if they can represent their surroundings completely or might be adapted prior to their use.

\subsection{Conflict-Aware Path Planning}
In this section, the impact of the classified \textit{conflicting} cells on a path planning algorithm is investigated. 

\begin{figure}[b]
\centering
\vspace{0.2cm}
 \raisebox{-0.2cm}[0pt][0pt]{
\hspace{0.1cm}
\begin{tikzpicture}
\begin{axis}[
axis line style={draw=none}, tick style={draw=none}, yticklabels={,,}, xticklabels={,,}, width=1.0\columnwidth, height=1.6cm, legend columns=4,
legend cell align={left},
legend style={/tikz/every even column/.append style={column sep=0.2cm}}
]
\addplot[color=red, opacity=0.0, forget plot]{0};
\addlegendimage{area legend, color=TAB_BLUE}
\addlegendentry{ego pose}
\addlegendimage{area legend, color=TAB_RED}
\addlegendentry{goal}
\addlegendimage{solid, color=TAB_BLUE}
\addlegendentry{path}
\addlegendimage{solid, color=TAB_ORANGE}
\addlegendentry{conflict. pose}
\end{axis}
\end{tikzpicture}}
  \subfloat[]{\label{fig:path_comparison}\frame{\includegraphics[trim=9cm 7.5cm 10cm 11cm, clip, width=0.55\linewidth]{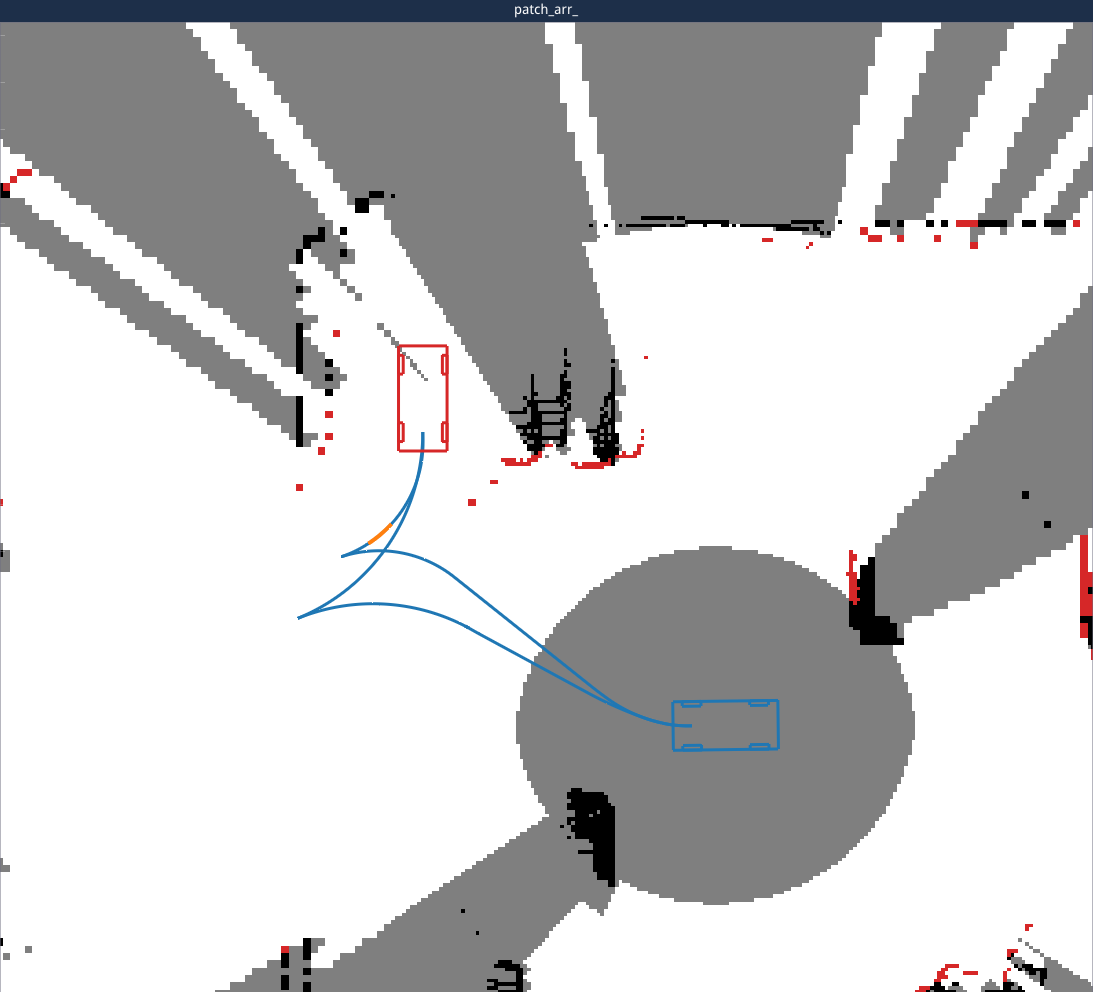}}}
  \hspace{0.5em}
  \subfloat[]{\label{fig:marked_parking_poses}\frame{\includegraphics[trim=9cm 10cm 10cm 10.4cm, clip, width=0.40\linewidth]{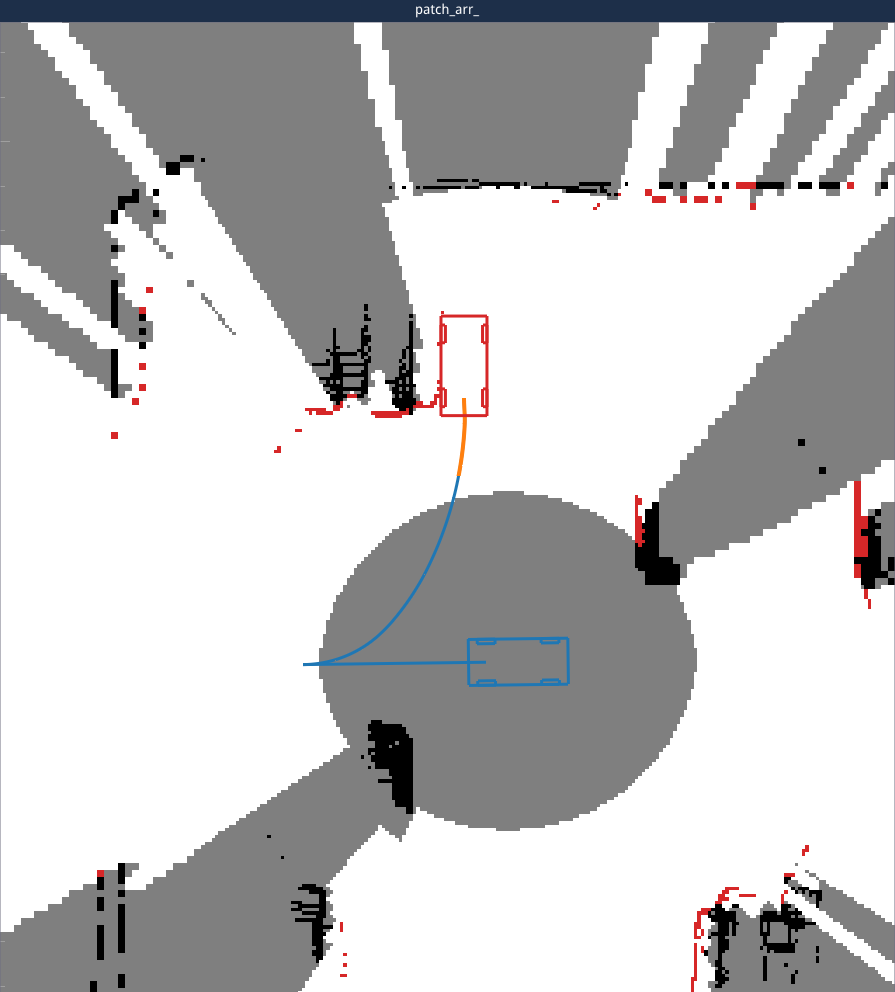}}}
  \caption{
  Planned paths on the categorized evidential grid map. The colors are equal to the ones in Fig.\ref{fig:colored_conflict}.
  (a) Intermediate poses are marked as conflicting. 
  If the cost of traversing \textit{conflicting} cells is increased, the planner finds a conflict-free path to the goal.
  (b) The goal and the last poses of the path are marked as conflicting. The goal is accepted. 
  The mentioned penalty makes no difference, as the goal itself is conflicting.
  }
\end{figure}
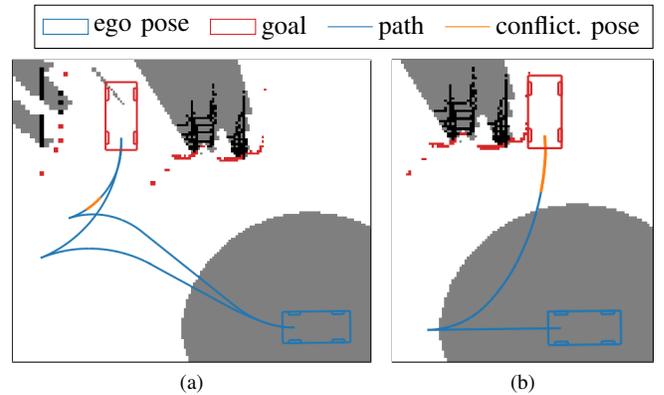
The first scene in Fig.~\ref{fig:path_comparison} shows two planned paths. Here, two paths without and with activated penalty for passing conflicted cells are depicted. Without the penalty, the path passes through conflicted areas. If conflicted cells are penalized, the path planning algorithm finds another path that takes a slight detour and avoids the \textit{conflicting} cells completely.

Fig.~\ref{fig:marked_parking_poses} shows a planned path to a goal that lies on conflicted cells. In this case, it cannot be avoided by the introduced penalty, which leads to \textit{conflicting} poses at the end of the path.

\begin{figure}[t]
\centering
\vspace{11mm}
 \raisebox{-0.2cm}[0pt][0pt]{%
\begin{tikzpicture}
\begin{axis}[axis line style={draw=none}, tick style={draw=none}, yticklabels={,,}, xticklabels={,,}, width=1.175\linewidth, height=1.7cm, legend columns=3,
legend cell align={left},
legend style={/tikz/every even column/.append style={column sep=0.3cm}}
]
\addplot[color=red, opacity=0.0, forget plot]{x};
\addlegendimage{area legend, color=TAB_BLUE}
\addlegendentry{ego pose}
\addlegendimage{area legend, color=TAB_RED}
\addlegendentry{goal}
\addlegendimage{area legend, color=TAB_ORANGE}
\addlegendentry{denied goal}
\addlegendimage{empty legend}
\addlegendentry{}
\addlegendimage{solid, color=TAB_BLUE}
\addlegendentry{path}
\addlegendimage{solid, color=TAB_ORANGE}
\addlegendentry{conflict. pose}
\end{axis}
\end{tikzpicture}}
  \subfloat[]{\label{fig:tunnel}\frame{\includegraphics[trim=14cm 5.5cm 3cm 6.5cm, clip, width=0.48\linewidth]{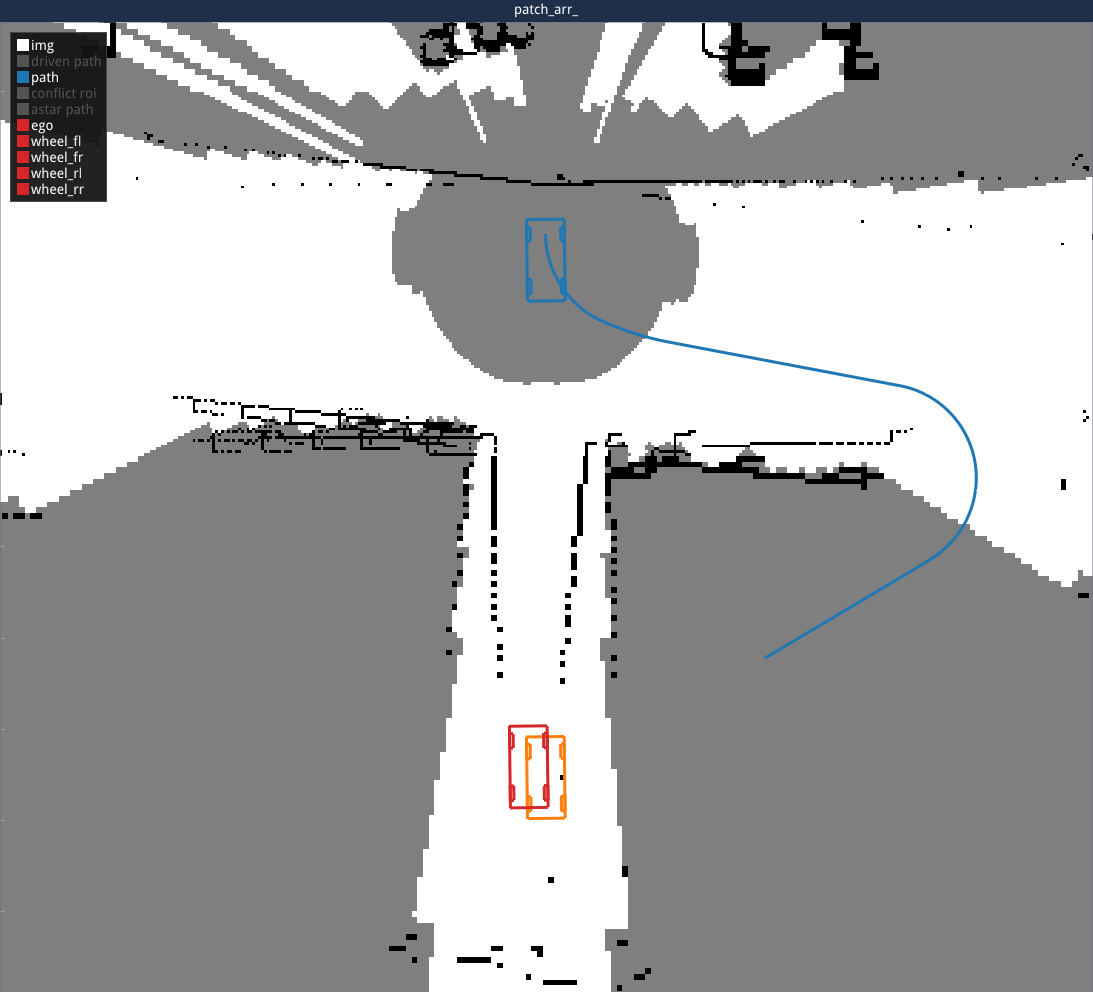}}}
  \hspace{0.5em}
    \subfloat[]{\label{fig:tunnel_conflict}\frame{\includegraphics[trim=8.5cm 4.5cm 8.1cm 7cm, clip, width=0.48\linewidth]{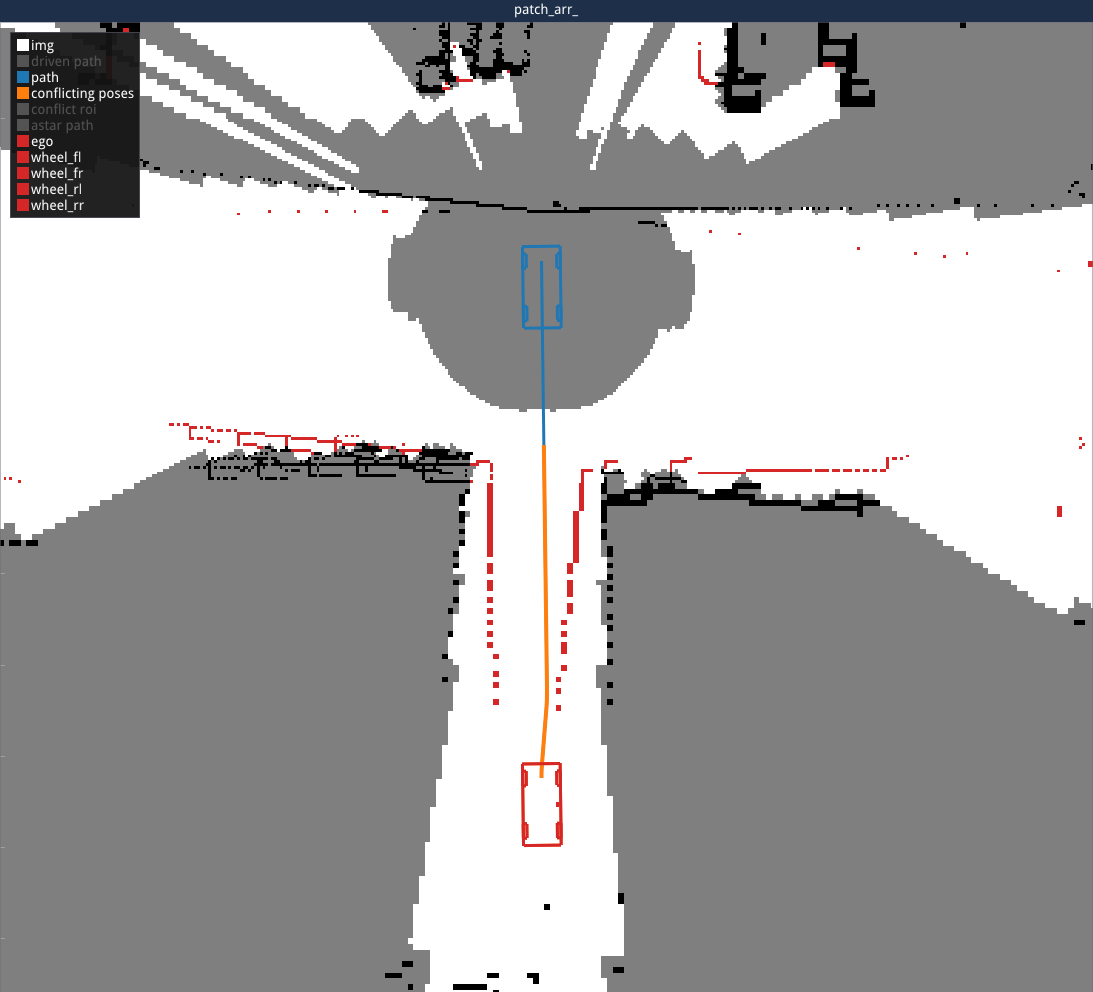}}}
    
  \subfloat[]{\label{fig:tunnel_inter}\frame{\includegraphics[trim=10cm 12cm 10cm 7cm, clip, width=0.48\linewidth]{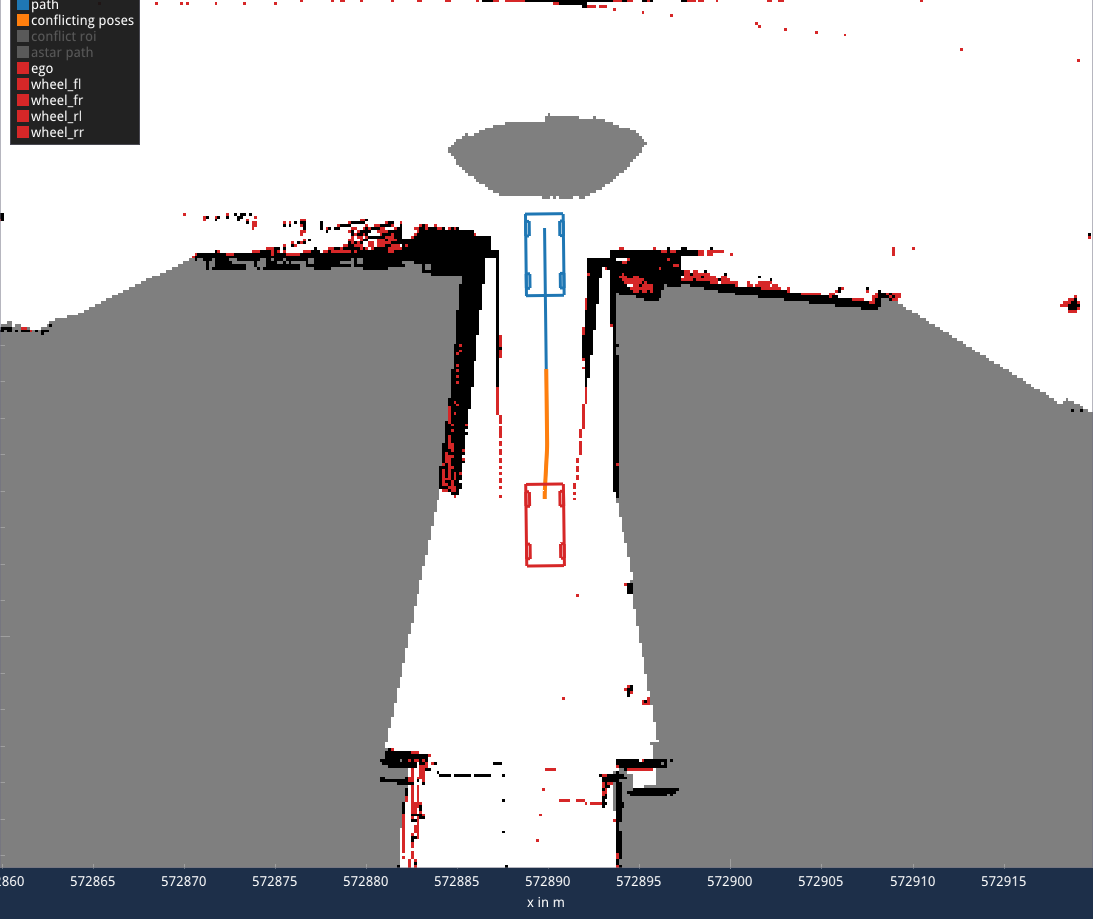}}}
  \hspace{0.5em}
\subfloat[]{\label{fig:tunnel_last}\frame{\includegraphics[trim=11cm 12cm 11cm 8.5cm, clip, width=0.48\linewidth]{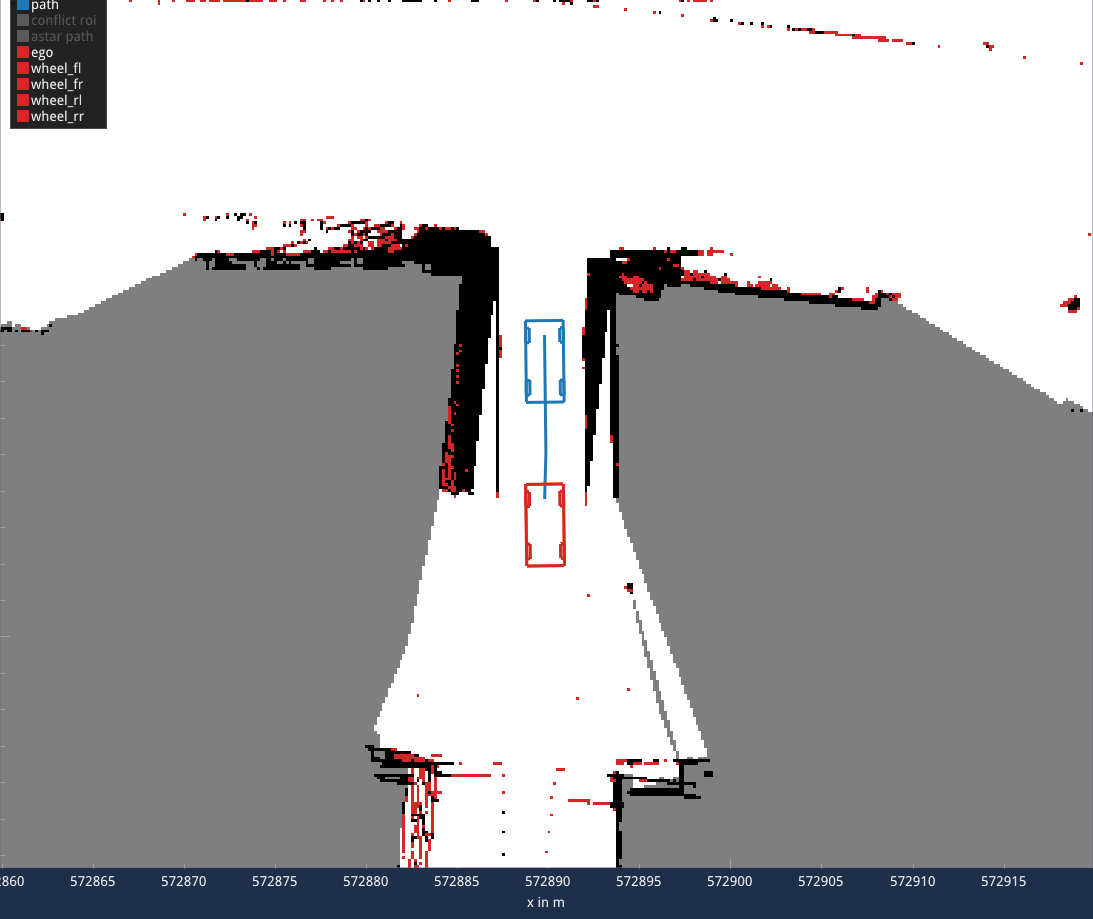}}}
  \caption{
  The ego vehicle must traverse a narrow passage despite having incorrectly calibrated LiDARs.
  (a) Path planning \emph{without} conflict handling. 
  The goal must be shifted to a valid one. Also, the passage is blocked and the planning algorithm tries to find a way around it. 
  (b) \emph{With} conflict handling, a collision-free path can be found, which must be traversed cautiously as it passes cells in conflict. 
  (c) The effect of the calibration error decreases. The number of \textit{conflicting} poses decreases. (d) The goal can be reached collision- and conflict-free.
  }
  \label{fig:tunnel_all}
\end{figure}
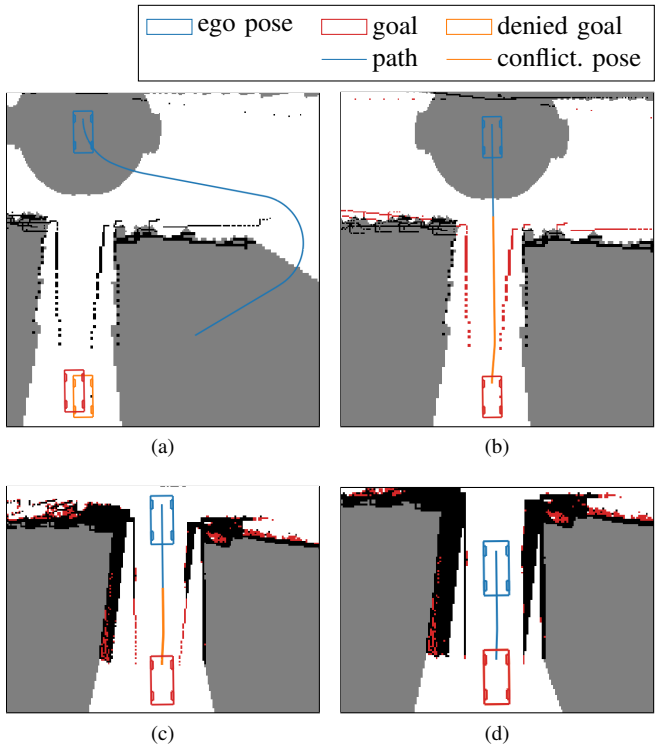

Fig.~\ref{fig:tunnel_all} visualizes a narrow passage that must be traversed by an automated vehicle to reach its goal. 
In real-world applications, this could be the road to the courtyard of a warehouse.
However, due to a calibration error, the LiDARs cannot measure the narrow passage correctly. The scene is further explained in the following and also shown in the video found at~\footnote{\href{https://youtu.be/94D2czuKpNw}{https://youtu.be/94D2czuKpNw}}

In the first scene in Fig.~\ref{fig:tunnel}, conventional grid mapping with conventional path planning is used. 
Here, the passage seems blocked, and the planning algorithm tries to find a path around the passage.

Fig.~\ref{fig:tunnel_conflict} depicts the same scene \emph{with} conflict handling. 
Here, the distance of the \textit{conflicting} cells to the ego vehicle is larger than the threshold $d\textsubscript{c}$. 
Hence, the path planning algorithm plans through them. 
However, the majority of the path is marked as in conflict.
\addtolength{\textheight}{-2.5cm} 

As the vehicle approaches the goal, entering the narrow passage in Fig.~\ref{fig:tunnel_inter}, the spatial impact of the calibration error decreases further. 
Here, half of the passage can be traversed without conflict. 
In the last scene depicted in Fig.~\ref {fig:tunnel_last}, the passage is now passed halfway. 
\textit{Conflicting} cells close to the vehicle are marked as \textit{occupied}. 
However, due to the decreased distance to the walls, the impact of the calibration error is so small that the passage is wide enough and the path planning algorithm can safely pass the narrow passage. Nevertheless, the conflict cannot be resolved completely which can be seen on the left wall of the passage. This is due to the sensors, which cannot reach the floor in close proximity and are therefore unable to classify as \textit{free}. Hence, the cells that were mapped as \textit{conflicting} or \textit{occupied} before are not overwritten with \textit{free}.
\begin{figure}[t]
\centering
\vspace{11mm}
 \raisebox{-0.2cm}[0pt][0pt]{%
\begin{tikzpicture}
\begin{axis}[axis line style={draw=none}, tick style={draw=none}, yticklabels={,,}, xticklabels={,,}, width=1.175\linewidth, height=1.7cm, legend columns=3,
legend cell align={left},
legend style={/tikz/every even column/.append style={column sep=0.3cm}}
]
\addplot[color=red, opacity=0.0, forget plot]{x};
\addlegendimage{area legend, color=TAB_BLUE}
\addlegendentry{ego pose}
\addlegendimage{area legend, color=TAB_RED}
\addlegendentry{goal}
\addlegendimage{area legend, color=TAB_ORANGE}
\addlegendentry{denied goal}
\addlegendimage{empty legend}
\addlegendentry{}
\addlegendimage{solid, color=TAB_BLUE}
\addlegendentry{path}
\addlegendimage{solid, color=TAB_ORANGE}
\addlegendentry{conflict. pose}
\end{axis}
\end{tikzpicture}}
  \subfloat[]{\label{fig:parking_conventional}\frame{\includegraphics[trim=13cm 14cm 11cm 3cm, clip, width=0.48\linewidth]{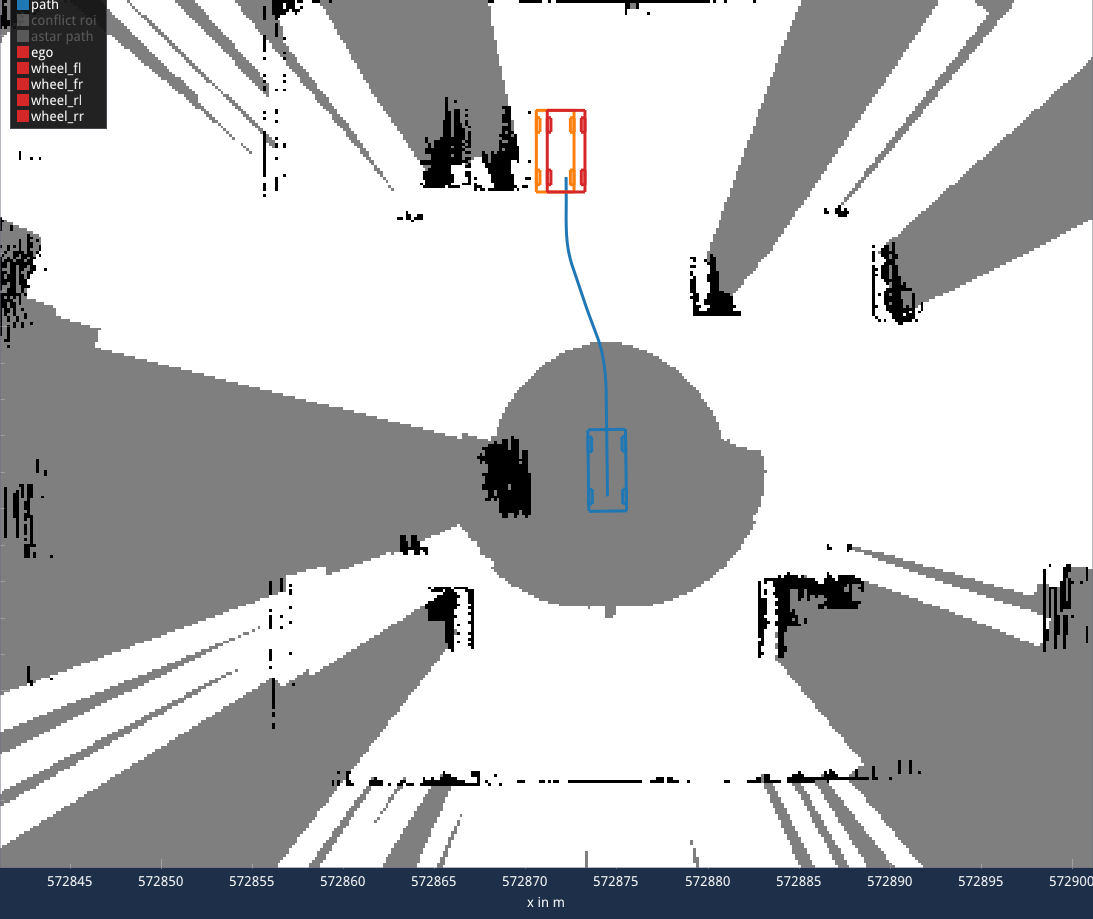}}}
  \hspace{0.5em}
    \subfloat[]{\label{fig:parking_start}\frame{\includegraphics[trim=13cm 12.5cm 11cm 4.5cm, clip, width=0.48\linewidth]{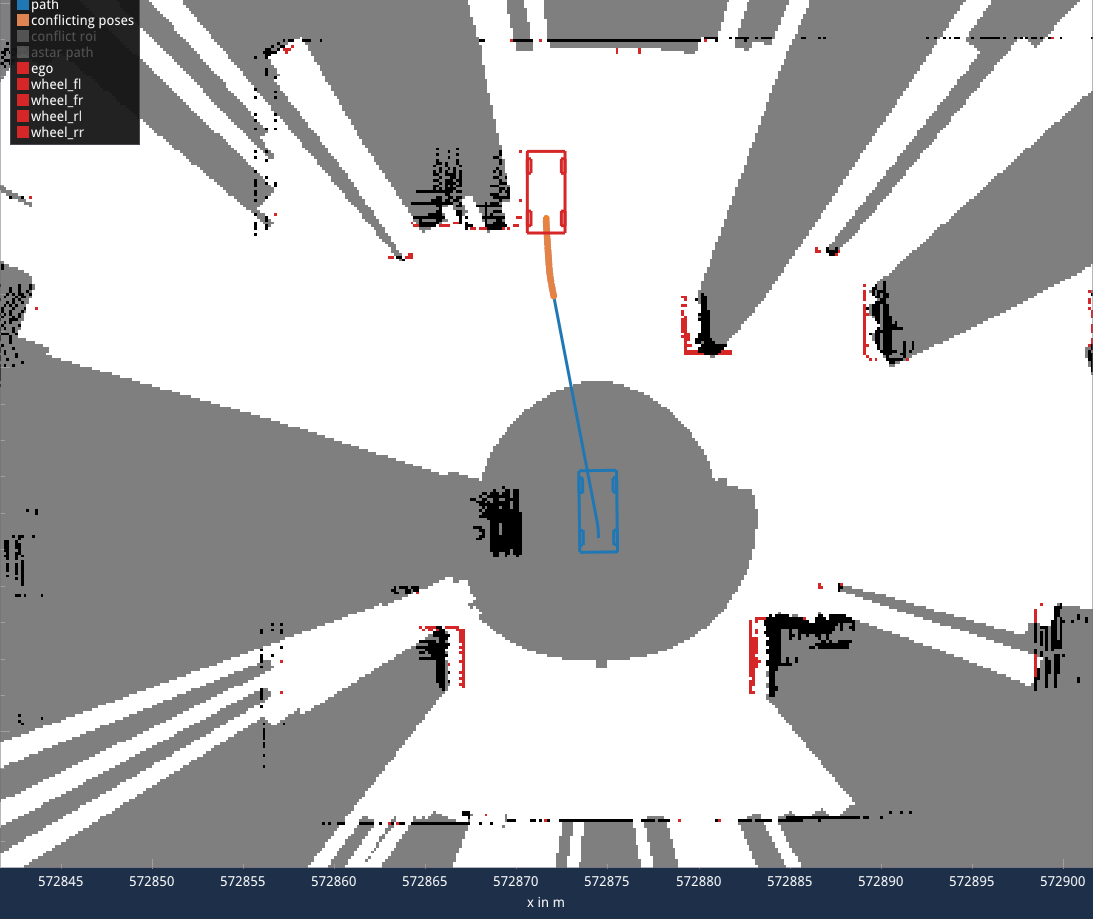}}}
    
  \subfloat[]{\label{fig:parking_inter}\frame{\includegraphics[trim=12cm 17.5cm 12cm 4cm, clip, width=0.48\linewidth]{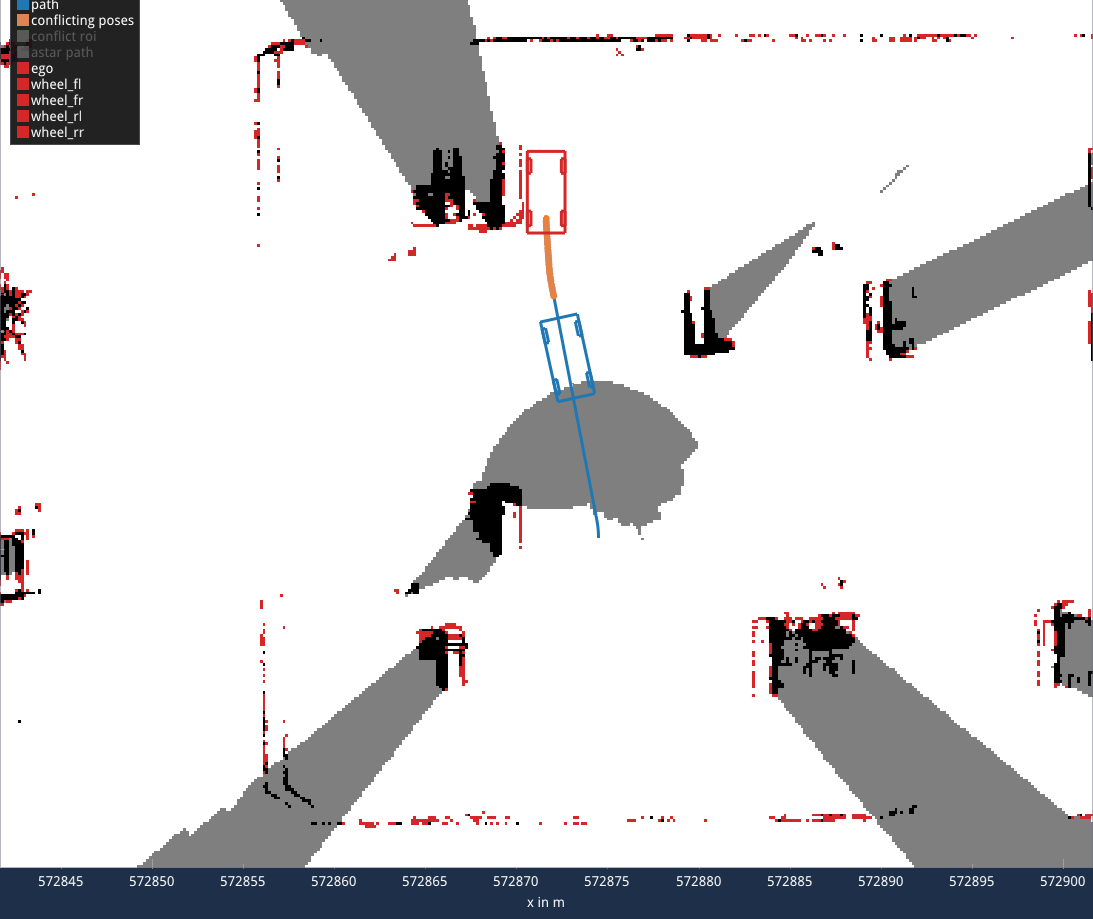}}}
  \hspace{0.5em}
    \subfloat[]{\label{fig:parking_fail_final}\frame{\includegraphics[trim=14cm 13cm 10cm 8.5cm, clip, width=0.48\linewidth]{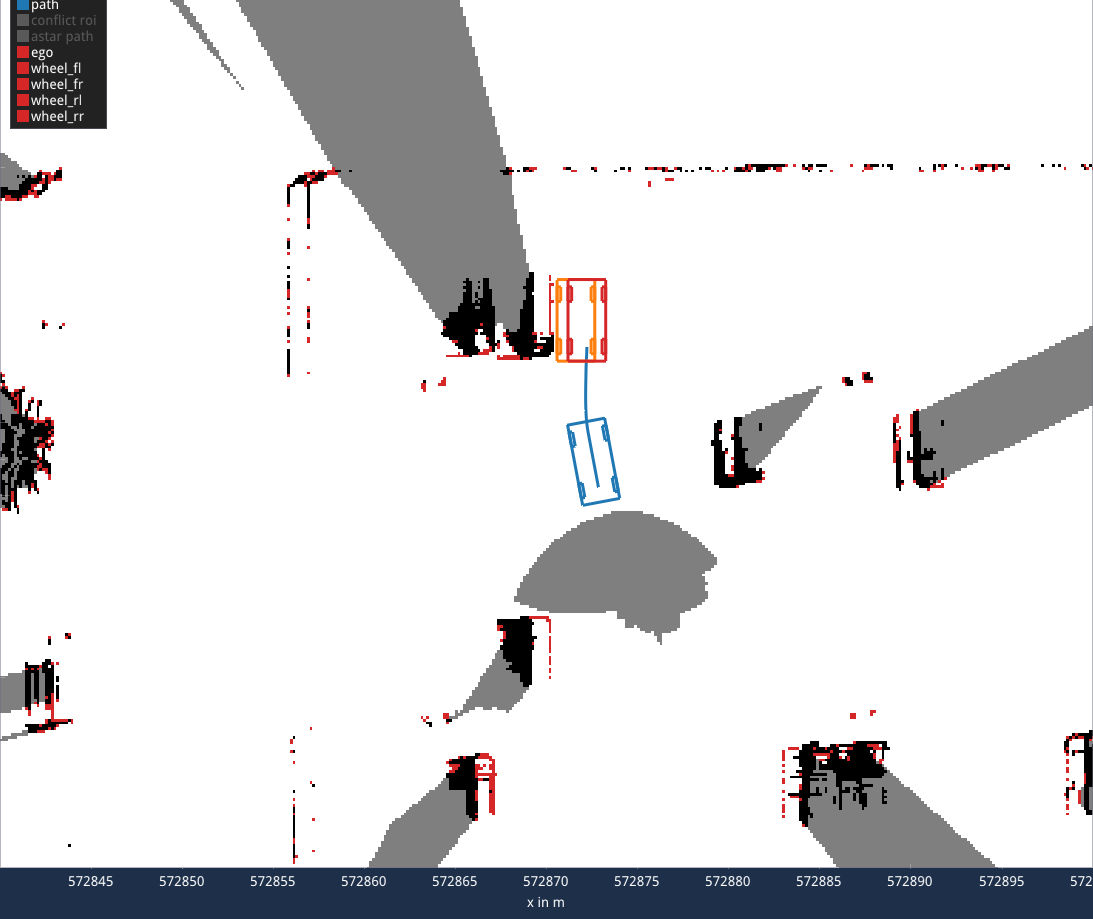}}}
  \caption{
  The vehicle has to park next to the cars on the top despite having incorrectly calibrated LiDARs.
  (a) Path planning \emph{without} conflict handling. 
  The goal must be shifted to a valid one. 
  (b) \emph{With} conflict handling, a collision-free path can be found.
  However, the goal and the last poses are conflicting. 
  (c) The conflict remains while approaching. 
  (d) \textit{conflicting} cells are set to occupied. 
  The goal is shifted to a collision-free one, leading to the same result as in (a).
  }
  \label{fig:parking}
\end{figure}
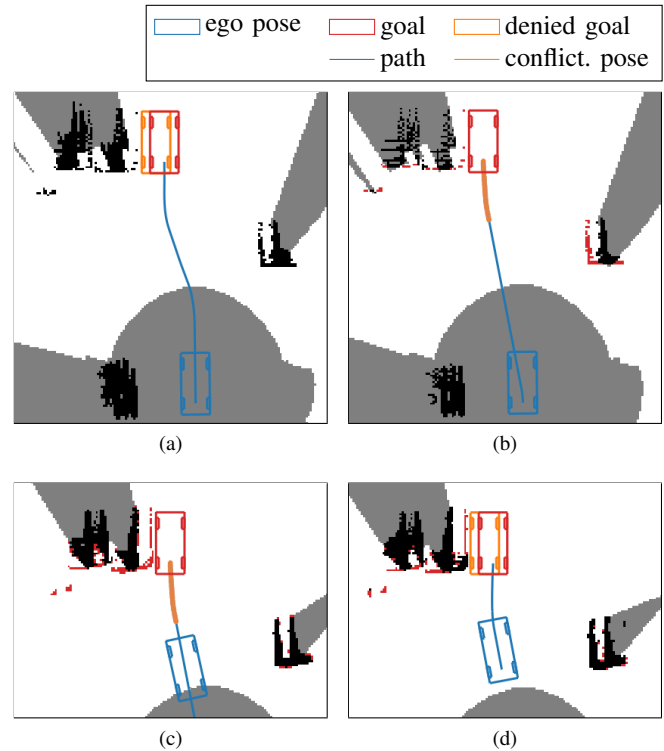

However, not all environmental sources of conflict have the property of decreasing their impact on approach. 
For example, a translational calibration error or some intrinsic errors like sensors with different modalities cannot be resolved. 
This is shown exemplarily in Fig.~\ref{fig:parking}.
Here, the goal must be shifted to a collision-free one in the proximity if conventional path planning is used.
If conflict-aware path planning is used, the vehicle tries to approach the goal, which is detected to lie on \textit{conflicting} cells. 
However, with decreasing distance, the conflicting cells remain, which leads to conflicted cells being marked as \textit{occupied}. 
As a result, the goal must be shifted as well which leads to the same result as when using conventional path planning in Fig.~\ref{fig:parking_conventional}.

Hence, the conflict-aware path planning approach can help to plan successful paths in degraded environment representations if the conflict can be resolved on approach. If this is not the case, this strategy leads to the same result as when using conventional path planning algorithms.
\FloatBarrier
\section{Conclusion}\label{sec:conclusion}
In this paper, we illustrated the identification of conflicting measurements in evidential grid maps by categorizing the fused opinions using SL. 
These categorized cells are used in a self-assessment module to calculate an overall degradation score. This score can be used to detect errors in the sensor setup, such as rotational and translational calibration errors and insufficient sensor capabilities.

Further, the information gained by the conflict-aware classification of the evidential grid map is passed to the motion planning stage, for which a conflict-aware path planning approach was presented.
It has the ability to plan safe and robust paths in degraded environment representations while avoiding regions with conflicting sensor data and being able to plan through them if necessary. This reduces the rate of unnecessary replanning steps or rejected goals that do not seem to be collision-free in earlier planning steps.

In future work, we want to focus on generalizing the presented approach by passing the information on the conflicting regions not only to the path planning algorithm but also to the underlying trajectory planning algorithm to realize cautious and safe trajectories in the presented degraded environment representations. 



%



\bibliographystyle{./IEEEtran.bst} 
\bibliography{./IEEEabrv,./references.bib, manual_references.bib}

\end{document}